\documentclass[10pt,twocolumn,letterpaper]{article}

\usepackage[final]{cvpr}
\usepackage[dvipsnames]{xcolor}
\definecolor{cvprblue}{rgb}{0.21,0.49,0.74}
\usepackage[pagebackref,breaklinks,colorlinks,citecolor=cvprblue]{hyperref}
\usepackage{algorithm}
\usepackage{algpseudocode}
\usepackage{amsmath}
\usepackage{annotate-equations}
\definecolor{myorange}{HTML}{FDE8D4}
\definecolor{myblue}{HTML}{D4D4FF}
\definecolor{mygreen}{HTML}{D4FFD4}
\definecolor{mygred}{HTML}{F8D5D5}

\title{Stable Score Distillation for High-Quality 3D Generation}

\author{
Boshi Tang$^{1,2}$\thanks{Equal contribution.}~~\thanks{Work done during an internship at IDEA.}~~,
Jianan Wang$^1$\footnote[1]{}~~\thanks{Corresponding author.}~~,
Zhiyong Wu$^2$, Lei Zhang$^1$\\
$^1$International Digital Economy Academy (IDEA)\\
$^2$Tsinghua University\\
}

\begin{document}
\maketitle
\begin{abstract}
Although Score Distillation Sampling (SDS) has exhibited remarkable performance in conditional 3D content generation, a comprehensive understanding of its formulation is still lacking, hindering the development of 3D generation. In this work, we decompose SDS as a combination of three functional components, namely mode-seeking, mode-disengaging and variance-reducing terms, analyzing the properties of each. We show that problems such as over-smoothness and implausibility result from the intrinsic deficiency of the first two terms and propose a more advanced variance-reducing term than that introduced by SDS.
Based on the analysis, we propose a simple yet effective approach named Stable Score Distillation (SSD) which strategically orchestrates each term for high-quality 3D generation and can be readily incorporated to various 3D generation frameworks and 3D representations.
Extensive experiments validate the efficacy of our approach, demonstrating its ability to generate high-fidelity 3D content without succumbing to issues such as over-smoothness.
\end{abstract}    
\section{Introduction}
\label{sec:intro}

3D content creation plays a crucial role in shaping the human experience, serving practical purposes such as the real-world production of everyday objects and the construction of simulated environments for immersive applications such as video games, AR/VR, and more recently, for training agents to perform general tasks in robotics. However, traditional techniques for 3D content generation are both expensive and time-consuming, requiring skilled artists with extensive 3D
modeling knowledge. Recent advancements in generative modelling have sparked a surge of interest in improving the accessibility of 3D content creation, with the goal to make the process less arduous and to allow more people to participate in creating 3D content that reflects their personal experiences and aesthetic preferences.

3D generative modelling is inherently more complex than 2D modelling, requiring meticulous attention to view-consistent fine geometry and texture.  Despite the increased intricacy, 3D data is not as readily available as its 2D image counterparts, which have propelled recent advances in text-to-image generation~\citep{dalle2,imagen,latentdiffusion}. 
Even with recent efforts of Objaverse~\citep{objaverse}, the uncurated 3D data only amounts to 10 million instances, in sharp contrast to the vast 5 billion image-text pairs available \citep{laion5b}.
As a result, utilizing 2D supervision for 3D generation has emerged as a prominent research direction for text-to-3D generation.
Notably, Score Distillation Sampling (SDS)~\citep{poole2022dreamfusion,SJC} is proposed to optimize a 3D representation so that its renderings at arbitrary views are highly plausible as evaluated by a pre-trained text-to-image model, without requiring any 3D data. 
As presented in DreamFusion~\citep{poole2022dreamfusion}, SDS enables the production of intriguing 3D models given arbitrary text prompts, but the results tend to be over-smooth and implausible (e.g., floaters). Subsequent works build upon SDS and enhance the generation quality with improvements in training practices. This is most effectively achieved by adopting higher resolution training, utilizing larger batch sizes, and implementing a coarse-to-fine generation approach with mesh optimization for sharper generation details~\citep{lin2023magic3d,chen2023fantasia3d,wang2023prolificdreamer}. 
While SDS remains fundamental and ubiquitously used for 3D generation, its theoretical understanding remains obscured and requires further exploration.

In this paper, we make extensive efforts to resolve the issues above, our contributions can be summarized as follows:
\begin{itemize}
    \item We offer a comprehensive understanding of Score Distillation Sampling (SDS) by interpreting the noise residual term (referred to as the SDS estimator) as a combination of three functional components: mode-disengaging, mode-seeking and variance-reducing terms. Based on our analysis, we identify that the over-smoothness and implausibility problems in 3D generation arise from intrinsic deficiency of the first two terms. Moreover, we identify the main training signal to be the mode-disengaging term.
    \item We show that the variance reduction scheme introduced by SDS incurs scale and direction mismatch problems, and propose a more advanced one.
    \item We propose Stable Score Distillation (SSD), which utilizes a simple yet effective term-balancing scheme that significantly mitigates the fundamental deficiencies of SDS, particularly alleviating issues such as over-smoothness and implausibility.
    \item We establish meaningful connections between our analytical findings and prevalent observations and practices in optimization-based 3D generation, such as the adoption of a large CFG scale and the generation of over-smoothed and floater-abundant results. These connections provide valuable perspectives for enhancing the optimization process of 3D generation.
    \item Extensive experiments demonstrate that our proposed method significantly outperforms baselines and our improvement is compatible with existing 3D generation frameworks and representations. Our approach is capable of high-quality 3D generation with vibrant colors and fine details.
\end{itemize}

\begin{figure*}[t]
\centering
\includegraphics[width=\linewidth]{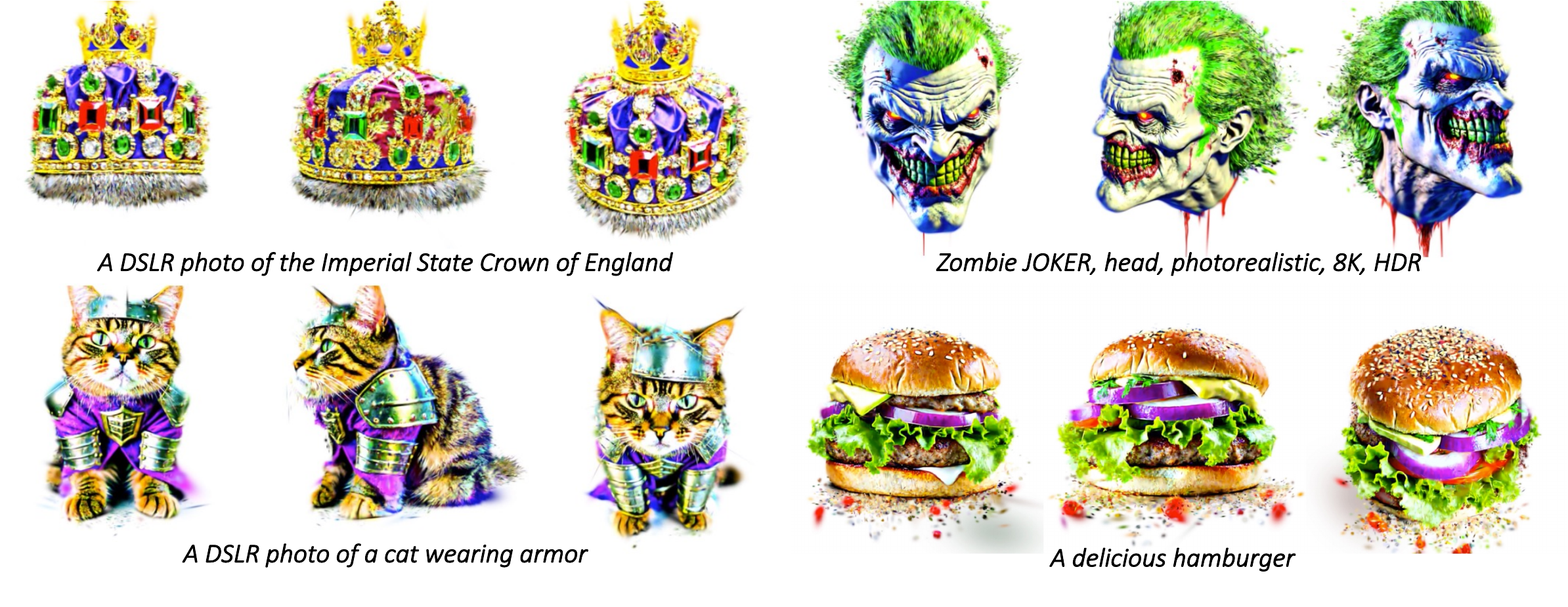}
\vspace{-2.2em}
\caption{3D Gaussian generation from text prompts.}
\label{fig:high_res_gs}
\vspace{-1.2em}
\end{figure*}

\begin{figure*}[t]
\centering
\includegraphics[width=\linewidth]{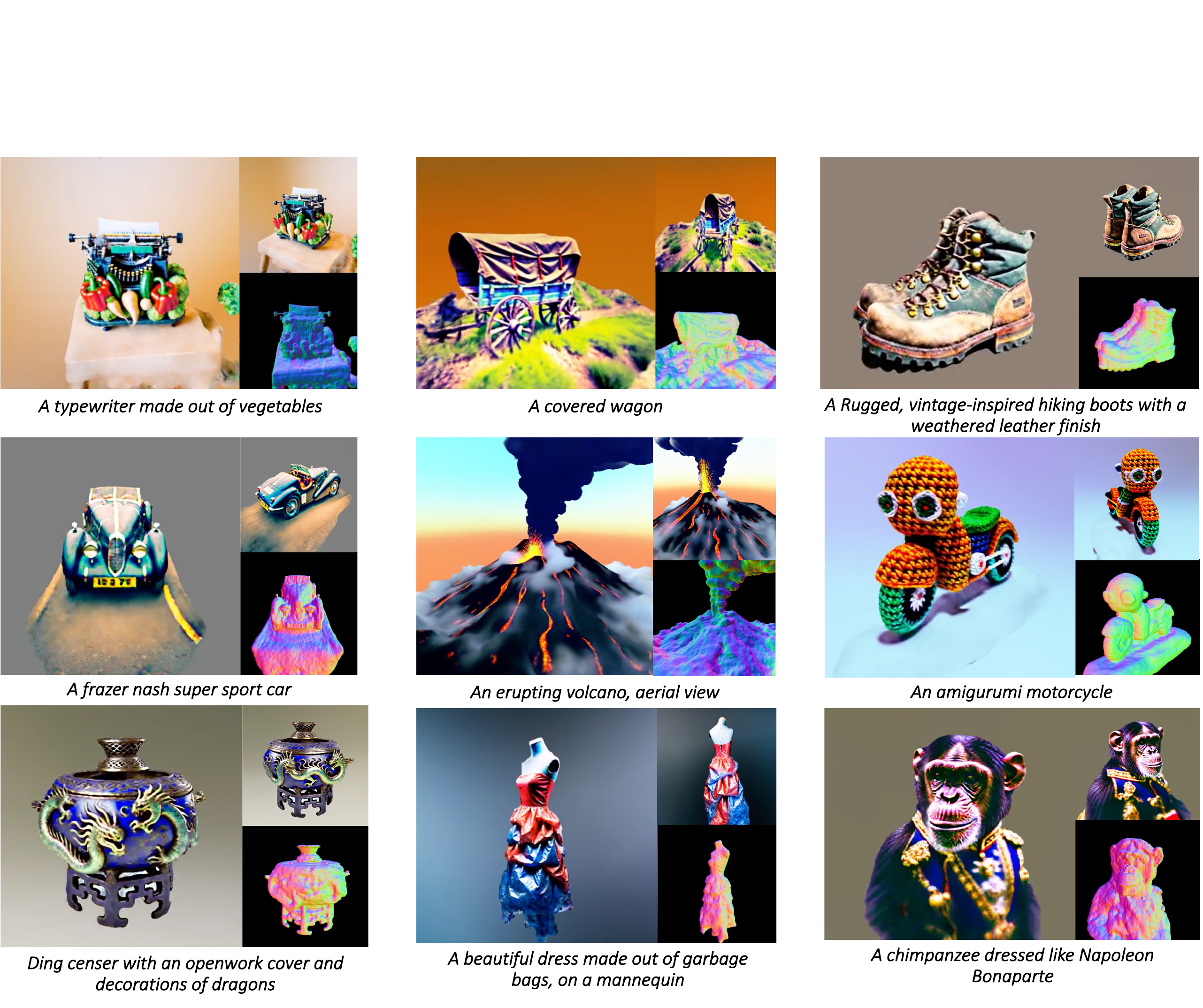}
\vspace{-2.2em}
\caption{NeRF generation from text prompts.}
\label{fig:high_res_nerf}
\vspace{-1.2em}
\end{figure*}

\begin{figure*}[t]
\centering
\includegraphics[width=\linewidth]{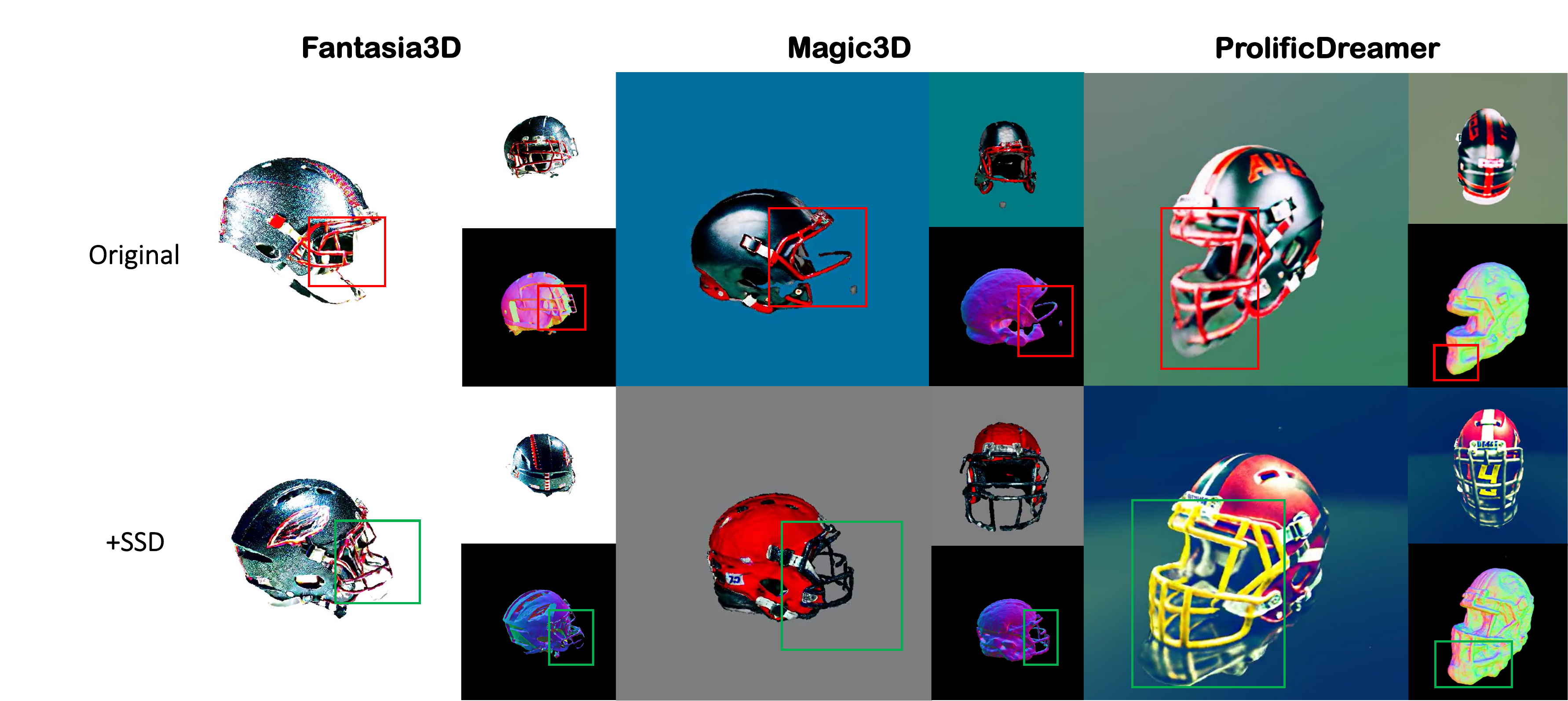}
\vspace{-2.2em}
\caption{Incorporating SSD to existing 3D generation frameworks consistently improves generation quality.}
\label{fig:ssd_incor}
\end{figure*}

\begin{figure*}[t]
\centering
\includegraphics[width=\linewidth]{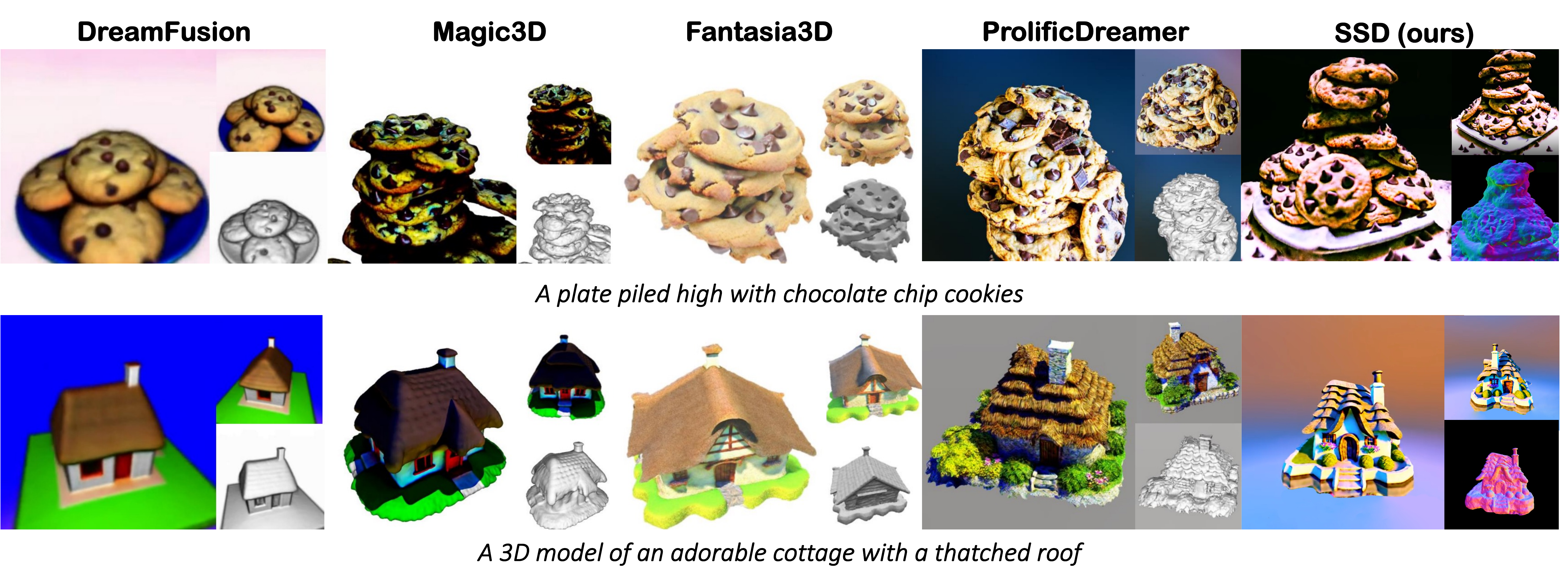}
\vspace{-2.2em}
\caption{Comparisons between SSD and SOTA methods on text-to-3D generation. Baseline results are obtained
from their papers.}
\label{fig:comp_baselines}
\vspace{-1.2em}
\end{figure*}

\begin{figure*}[t]
\centering
\includegraphics[width=\linewidth]{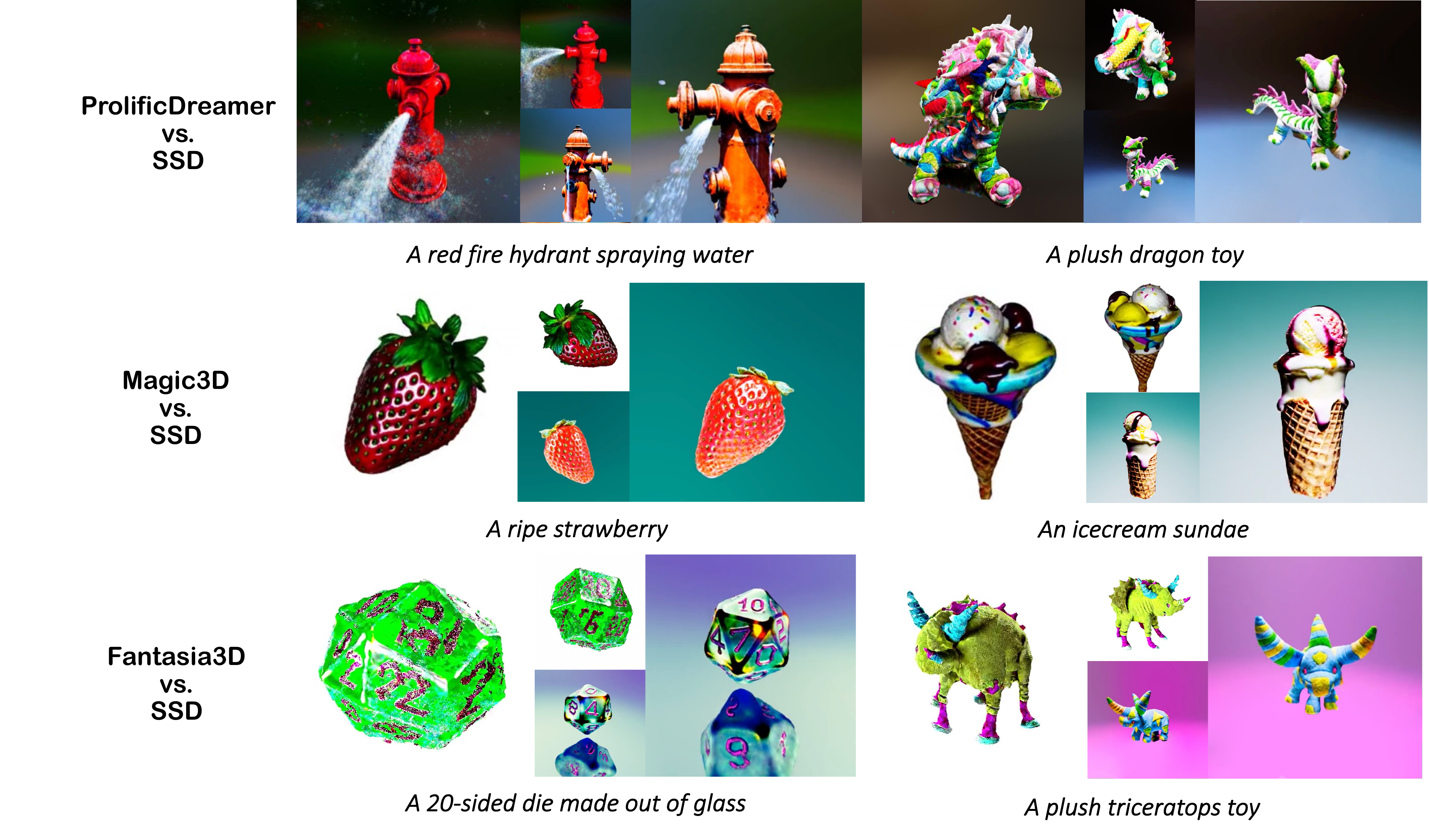}
\vspace{-2.2em}
\caption{More comparisons between SSD and SOTA methods on text-to-3D generation with more diverse prompts. For each text prompt, baseline results are obtained from theirs papers except for Fantasia3D, and presented on the left, while SSD results are shown on the right.  }
\label{fig:comp_baseline_diff_prompt}
\end{figure*}

\begin{figure*}[t]
\centering
\includegraphics[width=\linewidth]{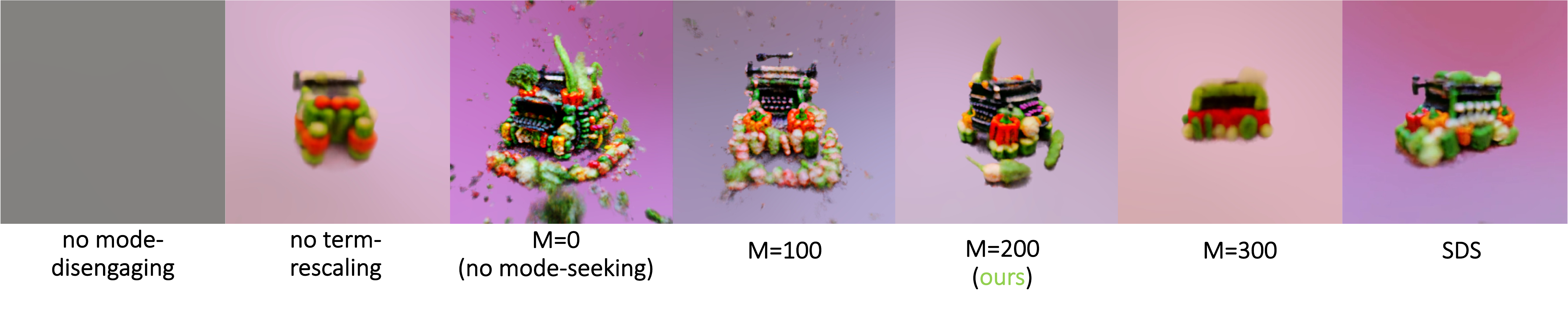}
\vspace{-2.2em}
\caption{Ablation study on the design choices of SSD. Better viewed when zoomed in.}
\label{fig:Ablation}
\vspace{-1.2em}
\end{figure*}
\section{Related Works}
\paragraph{Text-to-image Generation.}
Text-to-image models such as GLIDE~\citep{glide}, unCLIP~\citep{dalle2}, Imagen~\citep{imagen}, and Stable Diffusion~\citep{latentdiffusion} have demonstrated impressive performance in generating high-quality images.
The remarkable advancement is attributed to improvement in generative modeling techniques, particularly diffusion models~\citep{beatsgan, ddim, improved_ddpm}. The capability of generating diverse and creative images given arbitrary text is further enabled by the curation of large web datasets comprising billions of image-text pairs~\citep{laion5b, cc, cc12}. Recently, generating varied viewpoints of the same subject, notably novel view synthesis (NVS) from a single image has made significant progresses~\citep{liu2023zero, liu2023syncdreamer,shi2023mvdream, shi2023toss,weng2023consistent123,ye2023consistent} by fine-tuning pre-trained diffusion models on renderings of 3D assets, learning to condition the generation on camera viewpoints. NVS models can be readily applied to enhance image-to-3D generation, orthogonal to our improvements.
\paragraph{Diffusion-guided 3D Generation.} Recently, DreamFusion~\citep{poole2022dreamfusion} and SJC~\citep{SJC} propose to generate 3D content by optimizing a differentiable 3D representation so that its renderings at arbitrary viewpoints are deemed plausible by 2D diffusion priors. Such methods are commonly referred to as the optimization-based 3D generation. Subsequent works enhance the generation quality by utilizing 3D-aware diffusion models for image-to-3D generation~\citep{tang2023makeit3d, qian2023magic123, li2023sweetdreamer, long2023wonder3d}; adopting a coarse-to-fine generation approach with improved engineering practices such as higher-resolution training and mesh optimization~\citep{lin2023magic3d, chen2023fantasia3d}; exploring timestep scheduling~\citep{huang2023dreamtime}; or 
introducing additional generation priors~\citep{huang2023dreamwaltz, cao2023dreamavatar, jiang2023avatarcraft,liao2023tada,zhang2023avatarverse}. 
ProlificDreamer~\citep{wang2023prolificdreamer}, along with recent works of NFSD~\citep{katzir2023noise} and CSD~\citep{yu2023text} concurrent to ours provide new perspectives on SDS. however, there remains a gap in a comprehensive understanding of the SDS formulation, which is the focus of this work. 
\label{sec:related_work}
\section{Background}
In this section, we provide the necessary notations, as well as the background on optimization-based 3D generation including diffusion models, Classifier-Free Guidance (CFG)~\citep{ho2022classifier} and Score Distillation Sampling (SDS)~\citep{poole2022dreamfusion}. 

\subsection{Notations}
Throughout this paper, $\boldsymbol{x}$, or equivalently $\boldsymbol{x}_0$, is used to denote a natural image drawn from the distribution of images $\mathbb{P}(\boldsymbol{x})$. 
Given a 3D model parameterized by $\theta$, a volumetric renderer $g$ renders an image $g(\theta, c)$ according to $\theta$ and the camera pose $c$. $t$ denotes the diffusion timestep $\in(1, T)$.

\subsection{Diffusion Models}
\label{sec:diffusion_models}
Diffusion models assume a forward process where we gradually corrupt a sample by consecutively adding small amount of Gaussian noise in $T$ steps, producing a sequence of noisy samples $\boldsymbol{x}_1, \boldsymbol{x}_2, \dots, \boldsymbol{x}_T$, where $\mathbb{P}(\boldsymbol{x}_t | \boldsymbol{x}) = \mathcal{N}(\boldsymbol{x}_t ; \alpha_t \boldsymbol{x}, \sigma_t^2 \boldsymbol{I})$, whose mean and variance are controlled by pre-defined $\alpha_t$ and $\sigma_t$, respectively. 
A denoiser $\hat{\boldsymbol{\epsilon}_\phi}$ parameterized by $\phi$ is trained to perform the backward process by learning to denoise noisy images $\boldsymbol{x}_t=\alpha_t \boldsymbol{x} + \sigma_t \boldsymbol{\epsilon}$~\citep{ddpm}:
\begin{equation}
    \phi^* = \text{argmin}_\phi \mathbb{E}_{\boldsymbol{\epsilon}, t, y} [|| \hat{\boldsymbol{\epsilon}_\phi}(\boldsymbol{x}_t, t, y) - \boldsymbol{\epsilon} ||^2],
\end{equation}
where $\boldsymbol{\epsilon} \sim \mathcal{N}(\boldsymbol{0}, \boldsymbol{I})$, and $y$ is an user-defined condition which is usually a text prompt describing $\boldsymbol{x}$. After training, one can generate an image $\boldsymbol{x}' \sim \mathbb{P}(\boldsymbol{x}|y)$ by initiating with $\boldsymbol{x}_T  \sim \mathcal{N}(\boldsymbol{0}, \boldsymbol{I})$ and gradually denoise it with $\hat{\boldsymbol{\epsilon}_\phi}$.
Diffusion models can also be interpreted from the view of denoising score matching~\citep{song2020score}. Note that for each $t$, the noise term $\boldsymbol{\epsilon}=-\sigma_t \nabla_{\boldsymbol{x}_t}log \mathbb{P}(\boldsymbol{x}_t|\boldsymbol{x}, y, t)$. The property of denoising score matching~\citep{vincent2011connection} readily leads us to: 
\begin{equation}\label{eq:eps_as_score}
    \hat{\boldsymbol{\epsilon}_\phi}(\boldsymbol{x}_t, t, y) = -\sigma_t \nabla_{\boldsymbol{x}_t}log \mathbb{P}(\boldsymbol{x}_t ; y, t). 
\end{equation}
With this property, in our work we use $\mathbb{P}_\phi$ and $\mathbb{P}$ interchangeably.

\subsection{Classifier-Free Guidance (CFG)}
Classifier-Free Guidance (CFG)~\citep{ho2022classifier} trades off diversity for higher sample quality in the generation process of diffusion models. It has a hyperparameter called CFG scale, hereafter denoted as $\omega$, and works as follows,
\begin{equation*}
    \hat{\boldsymbol{\epsilon}_\phi}^{CFG}(\boldsymbol{x}_t, t, y) = (1+\omega)\cdot\hat{\boldsymbol{\epsilon}_\phi}(\boldsymbol{x}_t, t, y) - \omega\cdot \hat{\mathbf{\boldsymbol{\epsilon}}_\phi}(\boldsymbol{x}_t, t, \emptyset),
\end{equation*}
where $\emptyset$ represents the null symbol.

\subsection{Score Distillation Sampling (SDS)}\label{sec:van_sds}
SDS~\citep{poole2022dreamfusion} is an optimization-based 3D generation method that distills knowledge from pre-trained 2D diffusion models to 3D representations. It minimizes a weighted probability density distillation loss~\citep{oord2018parallel}, namely $\mathcal{L}_{SDS}(\phi, \boldsymbol{x}=g(\theta, c))=\mathbb{E}_{t}[(\frac{\sigma_t}{\alpha_t} w(t) \text{KL}(\mathbb{P}(\boldsymbol{x}_t | \boldsymbol{x}; y, t) || p_\phi(\boldsymbol{x}_t; y, t))]$:
\begin{equation}\label{eq:van_sds}
\nabla_\theta \mathcal{L}_{SDS}(\phi, \boldsymbol{x})=[w(t)(\colorbox{mygred}{$\hat{\boldsymbol{\epsilon}_\phi}^{CFG}(\boldsymbol{x}_t; y, t) - \boldsymbol{\epsilon}$}) \frac{\partial \boldsymbol{x}}{\partial \theta}],
\end{equation}
where $\boldsymbol{x}=g(\theta, c)$ and $w(t)$ re-scales the weights of gradients according to $t$, and $p_\phi(\boldsymbol{x}_t ; y, t)$ is the distribution of $\boldsymbol{x}_t$ implicitly represented by the score function $\hat{\boldsymbol{\epsilon}_\phi}$. DreamFusion~\citep{poole2022dreamfusion} introduces the $\boldsymbol{\epsilon}$ term for variance reduction. We call the resultant noise residual term highlighted in \colorbox{mygred}{red} the SDS estimator for short.
\section{Method}
\label{sec:method}
We start with problem setting in \cref{sec:prob_setting} and decompose the SDS estimator to three functional terms as shown in \cref{eq:sds_decompose}: the \colorbox{myorange}{\textit{mode-disengaging} term} (denoted as $h$ for its homogeneous form), the \colorbox{myblue}{\textit{mode-seeking} term} and the \colorbox{mygreen}{{}\textit{variance-reducing} term}. In \cref{sec:ana_mode_dis}, \cref{sec:ana_mode_seek}, and \cref{sec:ana_var_red} we analyze the mathematical properties, numerical characteristics and intrinsic limitations of each term, respectively.
Based on our analysis, we propose Stable Score Distillation (SSD) in \cref{sec:our_approach} which efficiently utilizes the distinctive properties of each term to augment each other for higher-quality 3D content generation.
\begin{equation}
\label{eq:sds_decompose}
\begin{split}
    \eqnmarkbox[red]{Psi2}{\hat{\boldsymbol{\epsilon}_\phi}^{CFG}(\boldsymbol{x}_t; y, t) - \boldsymbol{\epsilon}} &= \omega\cdot\eqnmarkbox[orange]{Psi2}{(\hat{\boldsymbol{\epsilon}_\phi}(\boldsymbol{x}_t, t, y) - \hat{\boldsymbol{\epsilon}_\phi}(\boldsymbol{x}_t, t, \emptyset))} \\
   &+ \eqnmarkbox[blue]{Psi2}{\hat{\boldsymbol{\epsilon}_\phi}(\boldsymbol{x}_t, t, y)} \\
   &- \eqnmarkbox[green]{Psi2}{\boldsymbol{\epsilon}}
\end{split}
\end{equation}

\subsection{Problem Setting}\label{sec:prob_setting}
We make a mild assumption that $\alpha_0=1, \sigma_0=0$; $\alpha_T=0, \sigma_T=1$ in our theoretical analysis for simplicity.\\
\textbf{3D Generation with 2D Supervision.}
The objective of 3D generation with 2D supervision is to create 3D assets whose renderings align well with the provided prompts. In other words, a necessary condition for successful 3D generation is that the renderings $g(\theta, c)$ of the trained 3D asset $\theta$ approach the modes of $\mathbb{P}(\boldsymbol{x};y)$. Note that this condition is necessary but not sufficient for 3D generation, lacking proper constraints for multi-view consistency.\\
\textbf{Mode Consistency Hypothesis.} Intuitively, an image that is plausible according to the conditional distribution, i.e. around a mode of $\mathbb{P}(\boldsymbol{x};y)$, should itself be a plausible image, namely around a mode of $\mathbb{P}(\boldsymbol{x})$. Therefore, we hypothesize that the modes of the conditional distribution $\mathcal{M}_{cond}$ form a subset of the modes of the unconditional distribution $\mathcal{M}_{uncond}$, formally $\mathcal{M}_{cond}\subseteq \mathcal{M}_{uncond}$. We refer to the modes in $\mathcal{M}_{uncond} - \mathcal{M}_{cond}$ as \textit{singular modes}.


\subsection{Analyzing the Mode-Disengaging Term}
\label{sec:ana_mode_dis}
\subsubsection{Mathematical Properties}
Substituting \cref{eq:eps_as_score} into the formulation of $h$, we obtain $h=-\sigma_t(\nabla_{\boldsymbol{x}_t} \text{log} \mathbb{P}_\phi(\boldsymbol{x}_t;y, t) - \nabla_{\boldsymbol{x}_t} \text{log} \mathbb{P}_\phi(\boldsymbol{x}_t; t))$. When optimizing a 3D representation $\theta$, the sub-terms of $h$, namely $-\sigma_t\nabla_{\boldsymbol{x}_t} \text{log} \mathbb{P}_\phi(\boldsymbol{x}_t;y, t)$ and $\sigma_t\nabla_{\boldsymbol{x}_t} \text{log} \mathbb{P}_\phi(\boldsymbol{x}_t;y, \emptyset)$, respectively drives $\theta$ such that $\mathbb{E}_{\boldsymbol{\epsilon}}[\boldsymbol{x}_t]=\alpha_t \boldsymbol{x}$ will: a) approach the conditional modes of $\mathbb{P}_\phi(\boldsymbol{x}_t;y, t)$, and b) distance the unconditional modes of $\mathbb{P}_\phi(\boldsymbol{x}_t; t)$ which are independent of the conditional prompt (for this reason we call $h$ mode-disengaging). We emphasize property b) to be critical, as it guides $\theta$ to avoid the over-smoothing problem caused by trapping points and transient modes, a concern encountered by the mode-seeking term (see \cref{sec:ana_mode_seek} for definition and analysis). In \cref{fig:intuition} (left), we illustrate that the mode-disengaging term has the capability to navigate out of the trapping point of the Mode-Seeking term induced from its transient modes, thanks to the disengaging force that avoids unconditional modes. This capability is crucial for effective 3D generation. Considering that in \cref{eq:sds_decompose}, only the mode-disengaging and mode-seeking terms provide supervision signals (note that $\mathbb{E}_{\boldsymbol{\epsilon}} [\boldsymbol{\epsilon}]=\boldsymbol{0}$), our analysis positions $h$ as the main supervision signal in 3D generation. Our analysis provides theoretical explanation to a concurrent work, CSD~\cite{yu2023text}, which also claims $h$ to be the main supervision signal.

\subsubsection{Intrinsic Limitation}\label{subsec:int_prob}
\begin{figure*}
    \centering
    \includegraphics[width=0.7\linewidth]{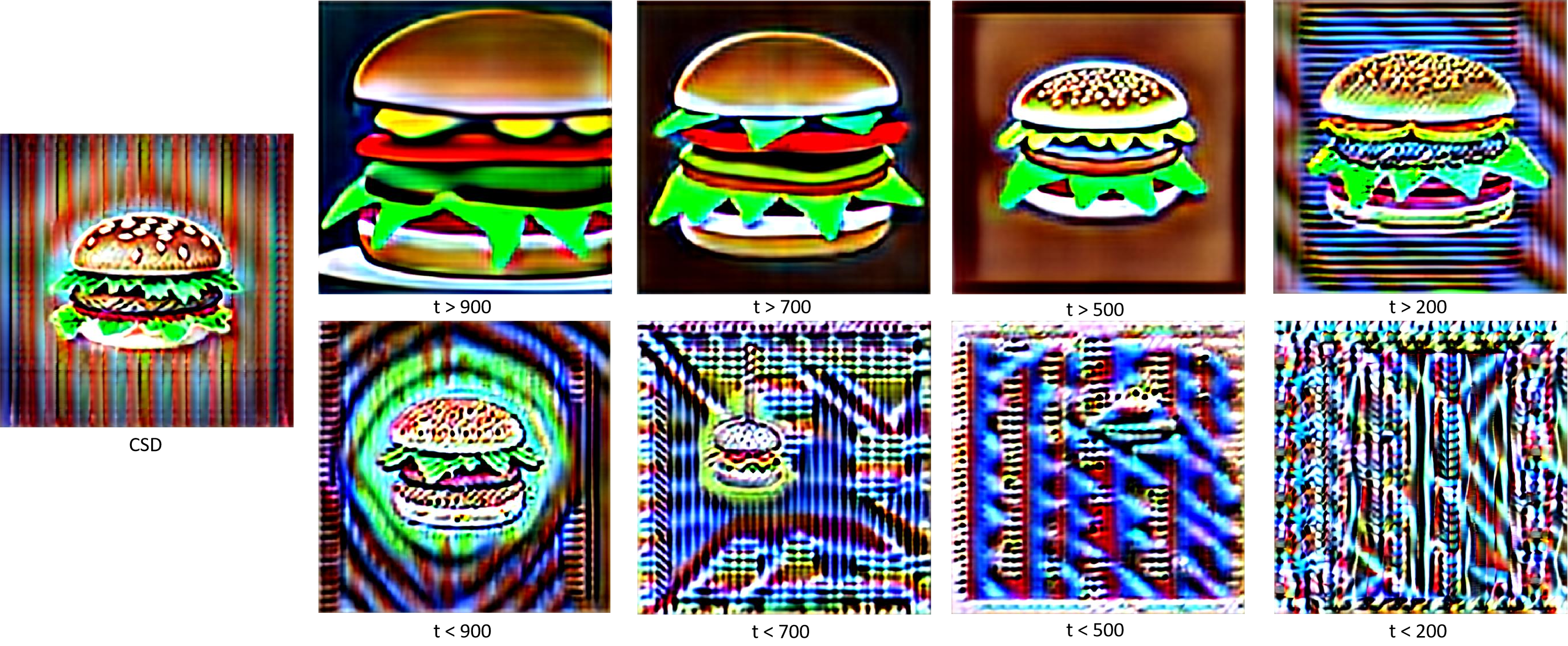}
    \caption{The correlation between $t$ and color-saturation problem of the \textit{mode-disengaging term}. The prompt is ``a hamburger''. Note that utilizing the mode-disengaging term alone with random $t \in(1, T)$ is equivalent to concurrent work of CSD~\citep{yu2023text}. $t>S$ means that the corresponding image is generated by sampling $t$ from $(S, T)$ and vice versa. It is evident that training signals from smaller $t$ contribute to the fineness of local features but also tend to produce implausible colors/textures. }
    \label{fig:t_200_color}
\end{figure*}

Based on previous discovery, one may be tempted to employ $h$ as the sole supervision signal in \cref{eq:van_sds} for 3D generation, akin to CSD~\citep{yu2023text}.
However, our experiments reveal a persistent issue of color-saturation in even 2D content generation when relying solely on $h$ (SSD, M=0 in \cref{fig:2d_comparison}). To delve deeper into this issue, we conduct experiments with varying values of $t$ and identify its particular association with small $t$, e.g., $t<200$ (see App. \cref{fig:t_200_color}). Note that as $t \rightarrow 0$, $\mathbb{P}_\phi (\boldsymbol{x}_t;y, t) \rightarrow \mathbb{P}_\phi (\boldsymbol{x};y, t)$ and $\mathbb{P}_\phi (\boldsymbol{x}_t;\emptyset, t) \rightarrow \mathbb{P}_\phi (\boldsymbol{x};\emptyset, t)$. As the term $h$ seeks to maximize $\frac{\mathbb{P}_\phi(\boldsymbol{x}_t;y, t)}{\mathbb{P}_\phi(\boldsymbol{x}_t;\emptyset, t)}$, when $t \rightarrow 0$, the learning target of $h$ gradually approaches maximizing $\frac{\mathbb{P}_\phi(\boldsymbol{x};y)}{\mathbb{P}_\phi(\boldsymbol{x})}$. However, the maximizing points for this ratio are often not around any modes of $\mathbb{P}_\phi(\boldsymbol{x};y)$, where $\mathbb{P}_\phi(\boldsymbol{x})$ is also high according to the mode consistency hypothesis. Worse still, they tend to be not around any singular modes of $\mathbb{P}_\phi(\boldsymbol{x})$, where $\mathbb{P}_\phi(\boldsymbol{x})$ is high and $\mathbb{P}_\phi(\boldsymbol{x};y)$ is low.
In other words, when $t \rightarrow 0$, the mode-disengaging term discourages $g(\theta, c)$ from converging to any mode of a natural image distribution $\mathbb{P}_{\phi}(\boldsymbol{x})$, let alone to modes of $\mathbb{P}_\phi(\boldsymbol{x};y)$, violating the necessary condition of 3D generation as presented in \cref{sec:prob_setting}. Therefore, special treatments are necessary to enable the 3D representation to converge to modes of $\mathbb{P}_{\phi}(\boldsymbol{x};y)$ when $t$ is small. We illustrate this problem in \cref{fig:intuition} (left), where the mode-disengaging trajectory converges to a point far from modes of $\mathbb{P}(\boldsymbol{x} ; y)$.


\subsection{Analyzing the Mode-Seeking Term}\label{sec:ana_mode_seek}
In light with the analysis in \cref{subsec:int_prob}, we aim to guide a 3D representation $\theta$ such that its rendering at arbitrary viewpoint seeks image distribution modes when $t$ is small. To achieve this, we employ $\hat{\boldsymbol{\epsilon}}_\phi(\boldsymbol{x}_t, y, t)$ as a viable choice, which points to the nearest mode of $\mathbb{P}_{\phi}(\boldsymbol{x}_t;y, t)$ according to \cref{eq:eps_as_score}, and conduct a thorough analysis of its properties.
\subsubsection{Numerical Properties}\label{sec:num_mode_seek}
\textbf{Scale.} Recall  $||\hat{\boldsymbol{\epsilon}}_\phi(\boldsymbol{x}_t, y, t)||=\sigma_t \cdot ||\nabla_{\boldsymbol{x}_t} \text{log} \mathbb{P}_\phi(\boldsymbol{x}_t;y, t)||$. Assuming $||\nabla_{\boldsymbol{x}_t} \text{log} \mathbb{P}_\phi(\boldsymbol{x}_t;y, t)||$ is bounded, one should expect $||\hat{\boldsymbol{\epsilon}}_\phi(\boldsymbol{x}_t, y, t)||$ to increase as $t$ increases. We validate this assumption numerically in Appendix \cref{fig:scale_mode_seeking}.\\
\textbf{Direction.}
As $t \rightarrow T$, $\nabla_{\boldsymbol{x}_t} \text{log} \mathbb{P}_\phi(\boldsymbol{x}_t;y, t)$ gradually transforms from the score of conditional distribution of natural images, i.e., $\nabla_{\boldsymbol{x}} log \mathbb{P}_\phi (\boldsymbol{x};y)$, to $\nabla_{\boldsymbol{x}_T} \mathcal{N}(\boldsymbol{x}_T ; \boldsymbol{0}, \boldsymbol{I})$, which are independent of $\boldsymbol{\epsilon}$ and collinear to $\boldsymbol{\epsilon}$ respectively (proved in \cref{sec:proof_linear_cor}). Therefore the linear correlation between $\hat{\boldsymbol{\epsilon}}_\phi(\boldsymbol{x}_t, y, t)$ and $\boldsymbol{\epsilon}$ as defined in \cite{puccetti2022measuring} is expected to increase as $t$ increases. We validate this assumption in  \cref{fig:linear_cor}.
\subsubsection{Intrinsic Limitations}\label{sec:over_smoothing}

\textbf{Over-smoothness.} According to proof (A.4) in DreamFusion~\cite{poole2022dreamfusion}, the mode-seeking term drives $\mathbb{P}(\boldsymbol{x}_t | \boldsymbol{x}; y, t) = \mathcal{N}(\alpha_t \boldsymbol{x}, \sigma_t^2 I)$ to align with the high-density region of $p_\phi(\boldsymbol{x}_t; y, t)$. Naturally, this means that during the 3D optimization process, this term directs the mean of the above Gaussian distribution, namely $\alpha_t \boldsymbol{x}$, towards modes of $\mathbb{P}_{\phi}(\boldsymbol{x}_t ; y, t)$. When $t$ is small (e.g., $\leq 200$), this behaviour is precisely what we aim for since $\mathbb{P}_{\phi}(\boldsymbol{x}_t; y, t) \approx \mathbb{P}_{\phi}(\boldsymbol{x}; y, t)$ and the mode-seeking term guides $\theta$ such that its rendering $g(\theta, c)$ converges towards the modes of $\mathbb{P}_{\phi}(\boldsymbol{x}; y, t)$. However, for large values of $t$, the situation is more complicated. For clarity, let's consider two modes in $\mathbb{P}_{\phi}(\boldsymbol{x}; y, t)$ located at $\boldsymbol{o}_1$ and $\boldsymbol{o}_2$. Ideally, we would like $\boldsymbol{x}$ to converge around them. Equivalently, we would like $\alpha_t \boldsymbol{x}\approx \alpha_t \boldsymbol{o}_1$ or $\alpha_t \boldsymbol{x}\approx \alpha_t \boldsymbol{o}_2$ for any $t$. We call $\alpha_t \boldsymbol{o}_*$ as \textit{induced modes}. However, when $t$ gets high, $\alpha_t$ is rather low. And now the induced modes gets so close that they ''melt'' into a single mode which lies between the induced ones. We assume such a mode is $\alpha_t \boldsymbol{o}_{tr}$, which we call \textit{transient mode}. Trivially, this means $\boldsymbol{o}_{tr}$ lies between $\boldsymbol{o}_1$ and $\boldsymbol{o}_2$. Thus the mode-seeking term drives $\alpha_t \boldsymbol{x}$ to seek the transient modes $\alpha_t \boldsymbol{o}_{tr}$, instead of the induced ones as expected, when $t$ is large. In other words, the 3D rendering $\boldsymbol{x}$ optimizes for $\boldsymbol{x}\rightarrow \boldsymbol{o}_{tr}$. We call $\boldsymbol{o}_{tr}$ a \textit{trapping point} in $\mathbb{P}_{\phi}(\boldsymbol{x}; y)$. Since the timestep $t$ is randomly sampled during optimization, the overall optimization process frequently revisits large $t$ and gets $\boldsymbol{x}$ trapped at $\boldsymbol{o}_{tr}$. This results in over-smoothness as $\boldsymbol{x}$ now converges to a middle point of plausible image modes, trying to ``average'' the plausible contents. See \cref{fig:intuition} for the visualization of the mode-seeking term causing model parameters to be trapped by transient modes.

\textbf{Variance.} $\hat{\boldsymbol{\epsilon}}_\phi(\boldsymbol{x}_t, y, t)$ exhibits high variance as pointed out by DreamFusion~\citep{poole2022dreamfusion}. We attribute this issue to the high correlation between the estimator and $\boldsymbol{\epsilon}$ as discussed in \cref{sec:num_mode_seek}. To effectively employ $\hat{\boldsymbol{\epsilon}}_\phi(\boldsymbol{x}_t, y, t)$ for enforcing mode-seeking behaviour with small $t$, it is imperative to reduce the variance associated with this term.

\textbf{Connection with common observations and practices in 3D content generation.} We find that many observations and practices of various frameworks have non-trivial connections with the previous analysis. Interested ones can refer to \cref{sec:connection} for details.

\subsection{Analyzing the Variance-Reducing Term}\label{sec:ana_var_red}
\subsubsection{A Naive Approach}\label{sec:naive_approach}
\begin{figure}
    \centering
    \includegraphics[width=0.5\textwidth]{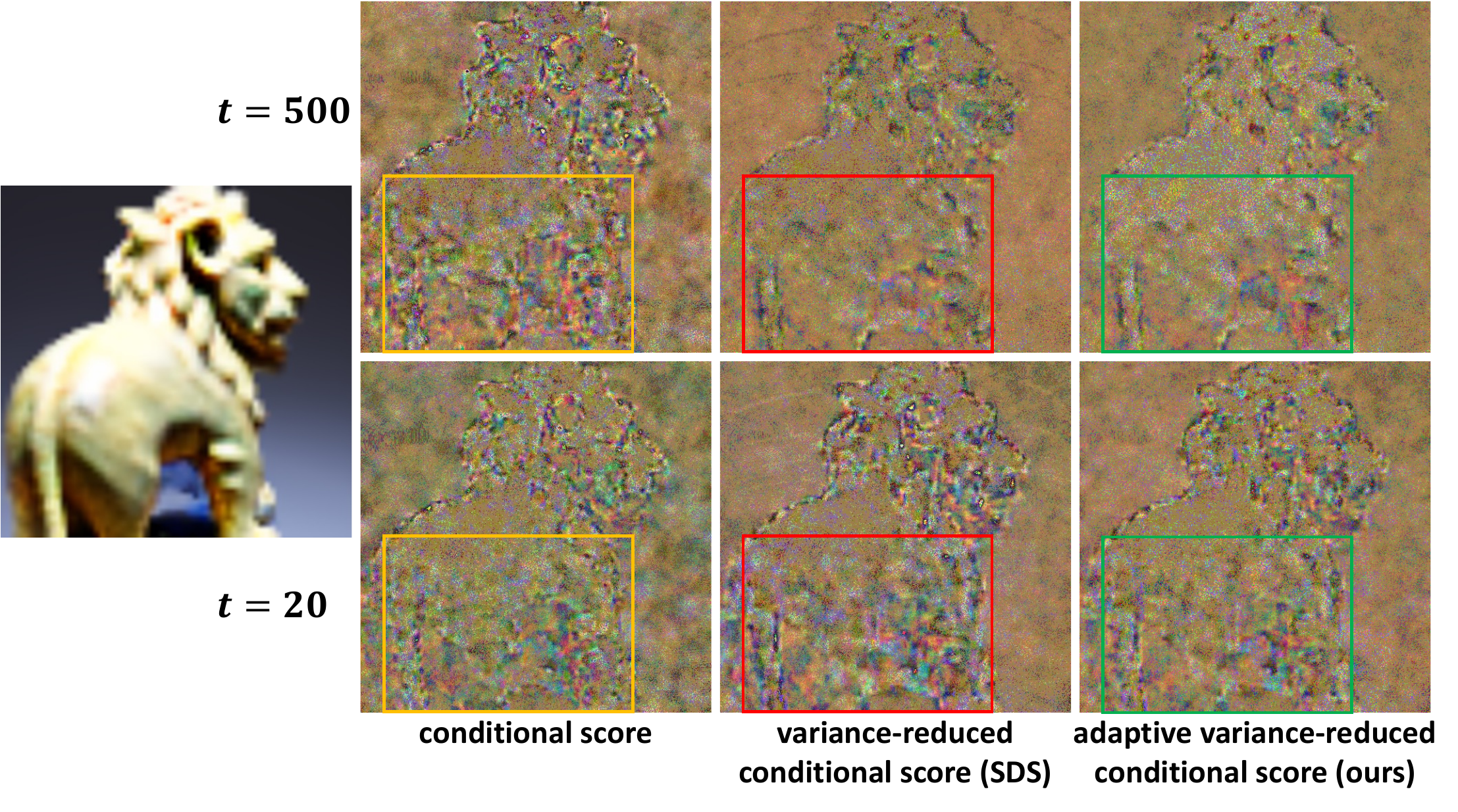}
    \vspace{0em}
    \caption{\textbf{Visualization of variance-reduction schemes for gradients.} Note that the SDS-like variance-reduction scheme does not effectively reduce variance of the conditional score for small $t$. When $t$ is large, e.g., 500, the adaptive variance-reduction term, $c$, is close to 1 and our variance reduction scheme performs similarly to SDS-like variance reduction scheme. On the other hand, when $t$ is small, our method significantly differs from SDS and produces smoother gradients. \textit{Conditional score} represents the gradients got by using the mode-seeking term only in optimization.} 
    \vspace{-1.2em}
    \label{fig:our_grad}
\end{figure}

Consider $\boldsymbol{b}=\hat{\boldsymbol{\epsilon}}_\phi(\boldsymbol{x}_t, y, t) - \boldsymbol{\epsilon}$, same as SDS with $\omega=0$. It may initially seem intuitive that $\hat{\boldsymbol{\epsilon}}_\phi(\boldsymbol{x}_t, y, t) \approx \boldsymbol{\epsilon}$ due to the denoising nature of training $\hat{\boldsymbol{\epsilon}}_\phi$ and thus $\boldsymbol{b}$ induces low variance, which is also an unbiased estimator for $\hat{\boldsymbol{\epsilon}}_\phi(\boldsymbol{x}_t, y, t)$ as $\mathbb{E}_{\boldsymbol{\epsilon}}[\boldsymbol{\epsilon}]=\boldsymbol{0}$. However, analysis from \cref{sec:ana_mode_seek} reveals that the scale and direction of the mode-seeking term are highly dependent on the diffusion timestep $t$ and thus $\boldsymbol{\epsilon}$ does not reduce variance always. For example, when $t=0$ where $\alpha_t=1$ and $\sigma_t=0$, we have $\hat{\boldsymbol{\epsilon}}_\phi=\boldsymbol{0}$ and the ``variance reduction'' term in $\boldsymbol{b}$ is actually the only source of variance! We illustrate this observation in \cref{fig:our_grad}. Consequently, it becomes important to reduce the variance of $\hat{\boldsymbol{\epsilon}}_\phi$ through a more meticulous design. 

\subsubsection{Adaptive Variance-Reduction}\label{sec:var_red_form}
We aim to reduce the variance of $\hat{\boldsymbol{\epsilon}}_\phi$ more efficiently by conditioning the variance-reduction term on $\boldsymbol{x}$ and $t$: $\hat{\boldsymbol{\epsilon}}_\phi - c(\boldsymbol{x}, t)\boldsymbol{\epsilon}$,
which remains an unbiased estimator of $\hat{\boldsymbol{\epsilon}}_\phi$ because the introduced coefficient $c(\boldsymbol{x}, t)$ is independent of $\boldsymbol{\epsilon}$, where $c$ is optimally chosen to maximally reduce the variance:
\begin{equation}
    \label{eq:c_min_var}
    c(\boldsymbol{x}, t)=\text{argmin}_k\ Var[\hat{\boldsymbol{\epsilon}}_\phi - k\boldsymbol{\epsilon}].
\end{equation}
Based on some mathematical treatments as detailed in \cref{sec:proof_close_c}, we instead optimize the objective function with $r$, which is a computationally cheap proxy for the optimal $c$ and defined as:
\begin{equation}
\label{eq:def_r}
    r(\boldsymbol{x}_t, \boldsymbol{\epsilon}, y, t)=\frac{\hat{\boldsymbol{\epsilon}}_\phi (\boldsymbol{x}_t,y,t)\cdot \boldsymbol{\epsilon}}{||\boldsymbol{\epsilon}||^2}.
\end{equation}
Formally, we define a variance-reduced mode-seeking estimator as:
\begin{equation}
\label{eq:r_red_var}
    \tilde{\boldsymbol{\epsilon}}_\phi (\boldsymbol{x}_t,y,t)=\hat{\boldsymbol{\epsilon}}_\phi (\boldsymbol{x}_t,y,t) - r(\boldsymbol{x}_t, \boldsymbol{\epsilon}, y, t) \cdot \boldsymbol{\epsilon}.
\end{equation}
It is worth noting that \cref{eq:r_red_var} is no longer an unbiased estimator of $\hat{\boldsymbol{\epsilon}}_\phi (\boldsymbol{x}_t,y,t)$ since $r$ is now dependent on $\boldsymbol{\epsilon}$. However, in practice, $r$ and $c$ are quite close to each other and the approximation with \cref{eq:r_red_var} does not result in any noticeable performance degradation. See \cref{fig:c_in_r} for numerical confirmation.

\subsection{Augmenting the Mode-Disengaging Term with Low-Variance Mode-Seeking Behaviour}\label{sec:our_approach}

We summarize the 3D generation supervision terms and their behaviour under different regimes of $t$ in Tab.~\ref{tab:terms_and_t_cases}. Evidently we should combine the merits of the mode-disengaging and the mode-seeking terms under different timestep regimes, which leads to our Stable Score Distillation (SSD) estimator. The design is quite simple: when $t>M$, where $M\geq 0$ is a pre-defined timestep threshold, we employ the mode-disengaging term for fast content formation. Otherwise we utilize the variance-reduced mode-seeking term as defined in \cref{eq:r_red_var} to guide the 3D model's renderings towards plausible image distribution. Notably, we further address the scale mismatch between the two cases by scaling the variance-reduced mode-seeking term to match the scale of the mode-disengaging one, inspired by the scale mismatch between the terms as shown in \cref{fig:norm_of_h_and_mode_seeking}. The detailed algorithm is provided in \cref{alg:ours}.

\begin{table}
\begin{tabular}{|c | c | c |} 
 \hline
 timestep $t$ & Mode-Disengaging & Mode-Seeking\\ [0.5ex] 
 \hline
 large & \textbf{trap escaping} & over-smoothness \\ 
 \hline
 small & implausibility (e.g., floaters) & \textbf{plausibility} \\
 \hline
\end{tabular}
\vspace{-0.8em}
\caption{Summarization of the supervision terms' behaviour under different $t$ regimes. The desired behaviours are emboldened.}
\label{tab:terms_and_t_cases}
\vspace{-1.6em}
\end{table}

\begin{algorithm}
\caption{Stable Score Distillation (SSD) Estimator}\label{alg:ours}
\begin{algorithmic}[1]
\State {\bfseries Input:} diffusion timestep $t$
\State {\bfseries Input:} noise $\boldsymbol{\epsilon}$ and corresponding noised rendering $\boldsymbol{x}_t$
\State {\bfseries Input:} generation condition, e.g., text prompt $y$
\State {\bfseries Input:} Timestep threshold $M$
\State {\bfseries Input:} pre-trained denoiser $\hat{\boldsymbol{\epsilon}}_\phi$, e.g., Stable Diffusion
\State {\bfseries Output:} SSD estimator to be used in place of SSD estimator as highlighted in \cref{eq:van_sds}
\State $h = \hat{\boldsymbol{\epsilon}}_\phi (\boldsymbol{x}_t, t, y) - \hat{\boldsymbol{\epsilon}}_\phi (\boldsymbol{x}_t, t, \emptyset)$
\If{$t > M$} \Comment{fast geometry formation}
    \State $E_{SSD}=h$
\ElsIf {$t \leq M$} \Comment{high-density chasing}
    \State Compute $r(\boldsymbol{x}_t, \boldsymbol{\epsilon}, y, t)$ according to \cref{eq:def_r}
    \State \textit{// variance-reduced score estimator}
    \State $\tilde{\boldsymbol{\epsilon}}_\phi (\boldsymbol{x}_t;y,t)=\hat{\boldsymbol{\epsilon}}_\phi (\boldsymbol{x}_t;y,t) - r(\boldsymbol{x}_t, \boldsymbol{\epsilon}, y, t) \cdot \boldsymbol{\epsilon}$
    \State \textit{// re-scaled score estimator}
    \State $E_{SSD}=\frac{||h||}{||\tilde{\boldsymbol{\epsilon}}_\phi (\boldsymbol{x}_t;y,t)||}\tilde{\boldsymbol{\epsilon}}_\phi (\boldsymbol{x}_t;y,t)$
\EndIf
    \State {\bfseries Return} $E_{SSD}$

\end{algorithmic}
\end{algorithm}


\vspace{-0.5em}

\section{Experiments}
\subsection{Implementation Details}\label{sec:imple_details_main_body}
Our experiments are mainly conducted with the \textit{threestudio} codebase~\cite{threestudio2023}. To save vram usage, during the first 5000 steps the training resolution is 64. The batch size for low-resolution and high-resolution stages are 8 and 1 respectively. We employ Stable Diffusion~\citep{latentdiffusion} v2.1-base as our score estimator $\hat{\boldsymbol{\epsilon}_\phi}$. We employ the ProgressBandHashGrid with 16 levels implemented in the codebase as our 3D representation. The learning rates are 0.01 for the encoding and 0.001 for the geometry networks. We conduct our experiments with one A100 80GB on Ubuntu 22.04LTS. Each asset is trained for 15000 steps. Also, during the first 5000 steps we sample $t$ from [20, 980] and anneal it to [20, 500] afterwards. The loss is SSD loss, plus the orient loss in the codebase with a coefficient 1000. Note that our method does not need to tune CFG scales like previous works~\cite{poole2022dreamfusion}.
\subsection{Evaluation on Numerical Experiment}
First, we demonstrate the efficacy of our SSD as a general-purpose estimator for mode approximation with a simple numerical experiment on a mixture-of-Gaussian distribution. Note that the mixture-of-Gaussian distribution is a general-purpose distribution approximator in that it can approximate any continuous distribution with any precision~\cite{GoodBengCour16}. Thus it suffices to evaluate our SSD on mixture-of-Gaussian. See \cref{sec:toy_example} for details.

\subsection{Evaluation on text-to-3D Generation}
\textbf{Comparison with baselines.}
We compare SSD with SOTA methods: DreamFusion~\cite{poole2022dreamfusion}, Fantasia3D~\cite{chen2023fantasia3d}, Magic3D~\cite{lin2023magic3d} and ProlificDreamer~\cite{wang2023prolificdreamer} on text-to-3D generation and provide the results in \cref{fig:comp_baselines} and \cref{fig:comp_baseline_diff_prompt}. SSD generates results that are more aligned with the prompts (e.g., the plate generation in the cookie example), plausible in geometry (e.g., the plush dragon toy example) and color (e.g., compared to Magic3D), and delicate.

\textbf{High-quality 3D Content Generation.} We evaluate the capability of SSD with diverse text prompts. As shown in \cref{fig:high_res_nerf} and \cref{fig:high_res_gs}, SSD is able to generate high-quality general 3D objects and diverse effects. With different 3D representations, it is highly efficient.

\begin{figure*}[t]
\centering
\includegraphics[width=0.8\linewidth]{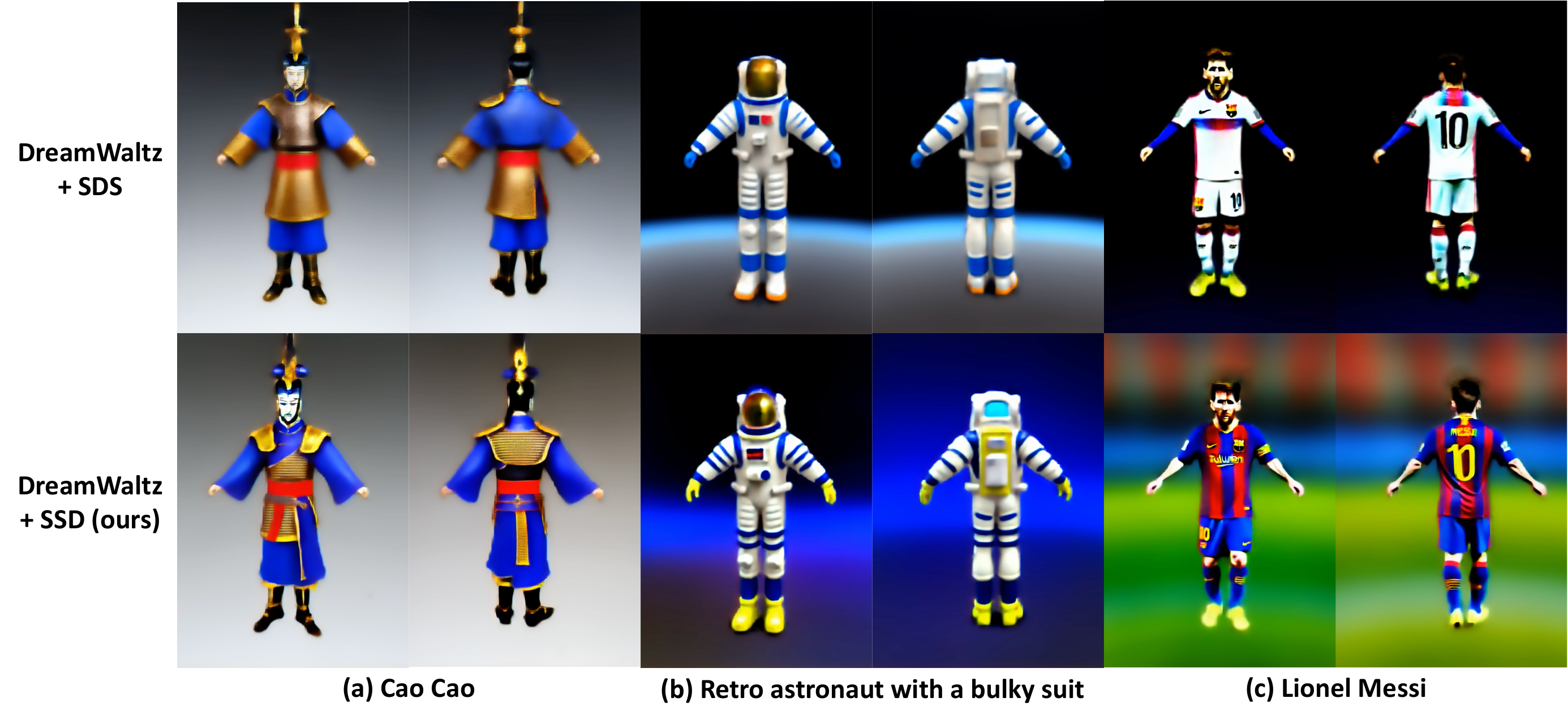}
\caption{Text-to-Avatar generation results using the DreamWaltz~\citep{huang2023dreamwaltz} framework, all with the same seed 0. Utilizing the proposed SSD instead of SDS improves both generation details (a-b) and overall plausibility (c). Note that SDS generates a Messi jersey with mixed styles in (c), indicative of a ``smoothing'' phenomenon as discussed in \cref{sec:over_smoothing}.
}
\label{fig:t2avatar}
\end{figure*}

\textbf{Compatibility with frameworks.} In \cref{fig:ssd_incor} and \cref{fig:t2avatar}, we show SSD's compatibility with existing 3D generation frameworks, no matter if they are general-purpose or specialized. SSD can be readily incorporated into them for quality improvement.

\textbf{User Studies.} We compare SSD to the four outlined baselines with user preference studies. Participants are presented with five videos side by side, all generated from the same text prompt. We ask the users to evaluate on two aspects: a) plausibility and b) details (existence of over-smoothness). We randomly select 100 prompts from the DreamFusion gallery for evaluation, with each prompt assessed by 10 different users, resulting in a total of 1,000 independent comparisons. The result reveals a preference of 3D models generated with our method over other baselines, $53.2\%$ for more generation plausibility and $61.4\%$  for finer details.
\subsection{Ablation Study}
\begin{figure*}
    \centering
    \includegraphics[width=0.75\linewidth]{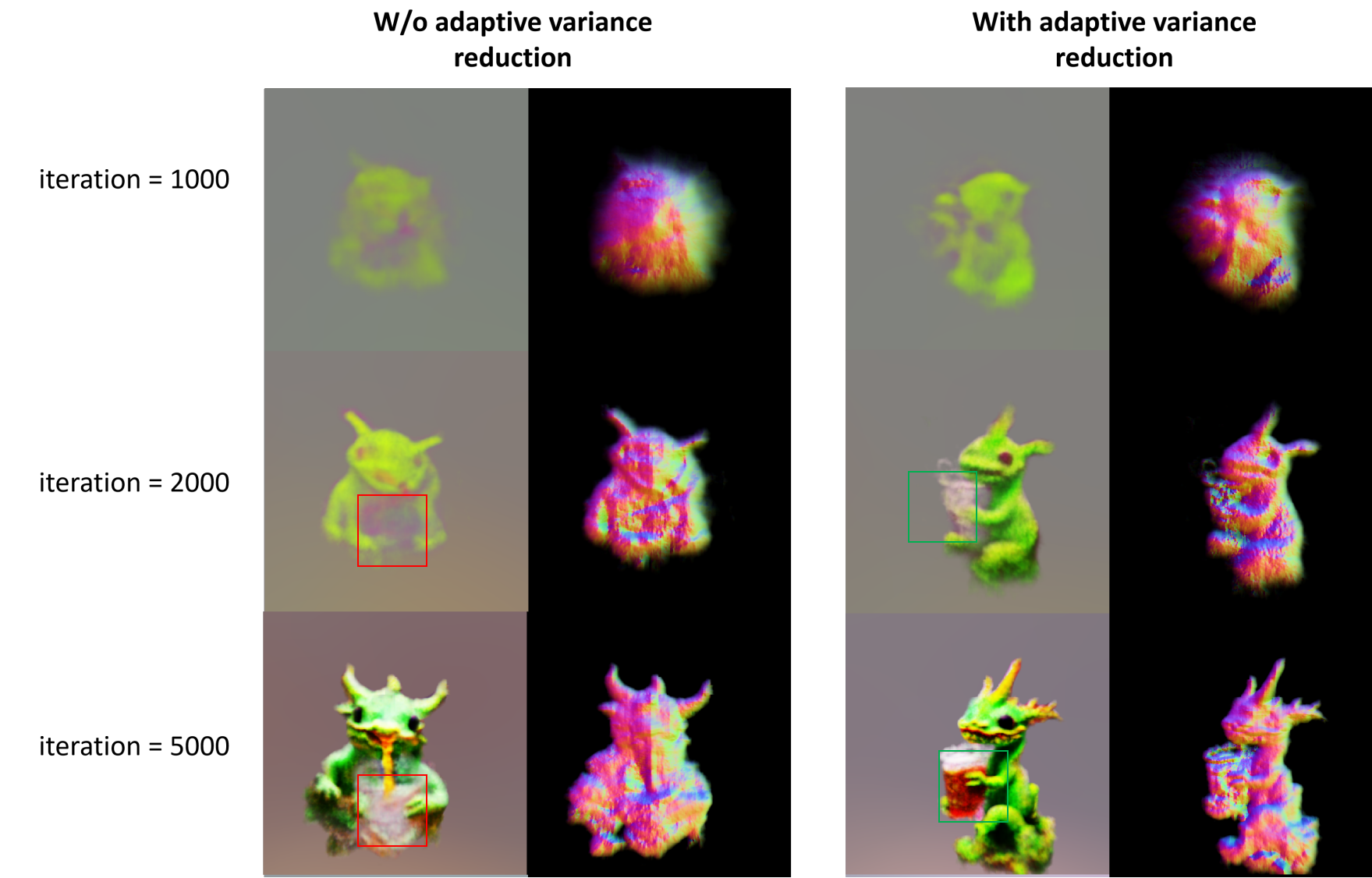}
    \caption{Comparison on prompt ``a baby dragon drinking boba''. With the proposed adaptive variance reduction scheme, SSD generates clear outline of the boba with only 2000 training steps. The accelerated learning pace on local features facilitates the successful generation of detailed boba, in contrast to the fuzziness observed in the absence of the adaptive variance reduction scheme.}
    \label{fig:adap_var_reduction_2}
\end{figure*}

\begin{figure*}
    \centering
    \includegraphics[width=0.75\linewidth]{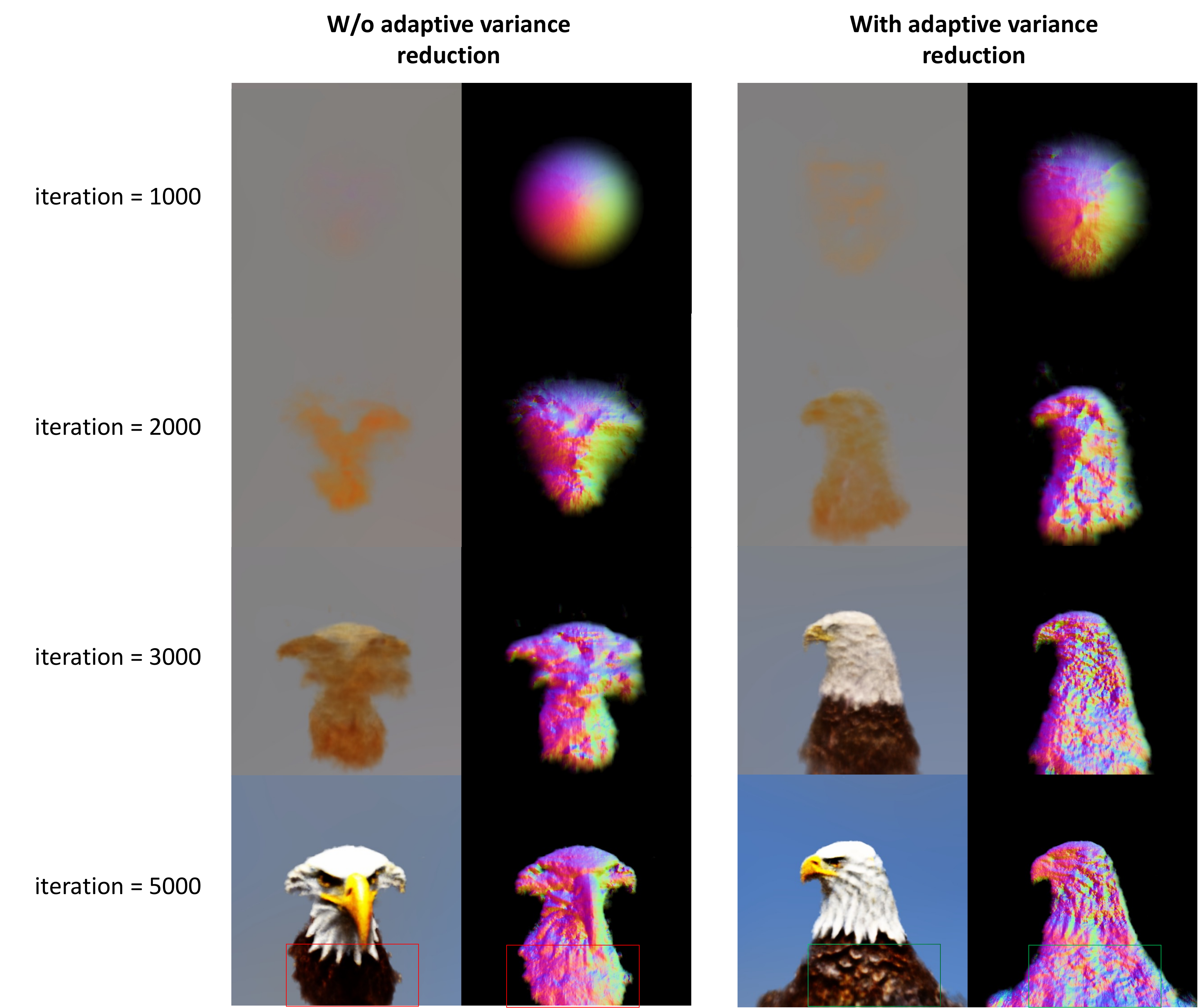}
    \caption{Comparison on prompt ``a bald eagle carved out of wood''. With the adaptive variance reduction scheme, SSD learns faster and generates plausible squama and colors of the carved eagle with only 3000 training steps. This scheme effectively alleviates the over-smoothing problem and the final generation shows delicate squama whereas the same region is overly smooth without the scheme.}
    \label{fig:adap_var_reduction_3}
\end{figure*}

In \cref{fig:Ablation}, \cref{fig:adap_var_reduction_2}, and \cref{fig:adap_var_reduction_3}, we assess all the designs of SSD. Removing of any of them can result to a general decrease in generation quality. Specifically, we see higher $M$ causes the result to be more smooth with fewer local details, as expected. The three terms, namely mode-seeking, mode-disengaging and variance reduction terms are all necessary for successful 3D generation. The variance reduction term can accelerate training process and help the emergence of fine details even in the early stage of training. Also, without the term rescaling mechanism, the training process suffers from the scale mismatch between mode-seeking and mode-disengaging terms, which has been pointed out by our analysis, and produces blurred 3D assets. Particularly, a naive combination of the three terms, like SDS, is inefficient and causes blurs.

\vspace{-0.5em}
\section{Conclusion}
We propose Stable Score Distillation (SSD) for high-quality 3D content generation, based on a comprehensive understanding of the widely used score distillation sampling (SDS). We interpret SDS as a combination of mode-disengaging, mode-seeking and variance-reducing terms, analyzing their distinct properties. This interpretation allows us to harness each term to its fullest potential and to leverage their complementary nature. Our analysis establishes rich connections to prevalent observations and practices in 3D content generation.
Extensive experiments demonstrate the effectiveness of our approach for generating high-fidelity 3D content without succumbing to issues such as over-smoothness and severe floaters.
\clearpage
{
    \small
    \bibliographystyle{ieeenat_fullname}
    \bibliography{main}
}
\clearpage
\clearpage
\setcounter{page}{1}
\maketitlesupplementary

In this supplementary material, \cref{appx:more_results} presents more qualitative results of SSD on diverse prompts. We provide proofs in \cref{sec:proof_close_c}, \cref{sec:subsec_for_proof_c} and \cref{sec:proof_linear_cor}, as well as numerical studies in \cref{sec:numerical_exp} to support our method's assumptions and analysis. \cref{sec:connection} shows the connection between prevalent observations and our analysis. \cref{sec:toy_example} gives a numerical experiment that shows the efficiency of SSD as a general-purpose mode approximator.

\section{More Qualitative Results}\label{appx:more_results}

\begin{figure*}[t]
\centering
\includegraphics[width=0.8\linewidth]{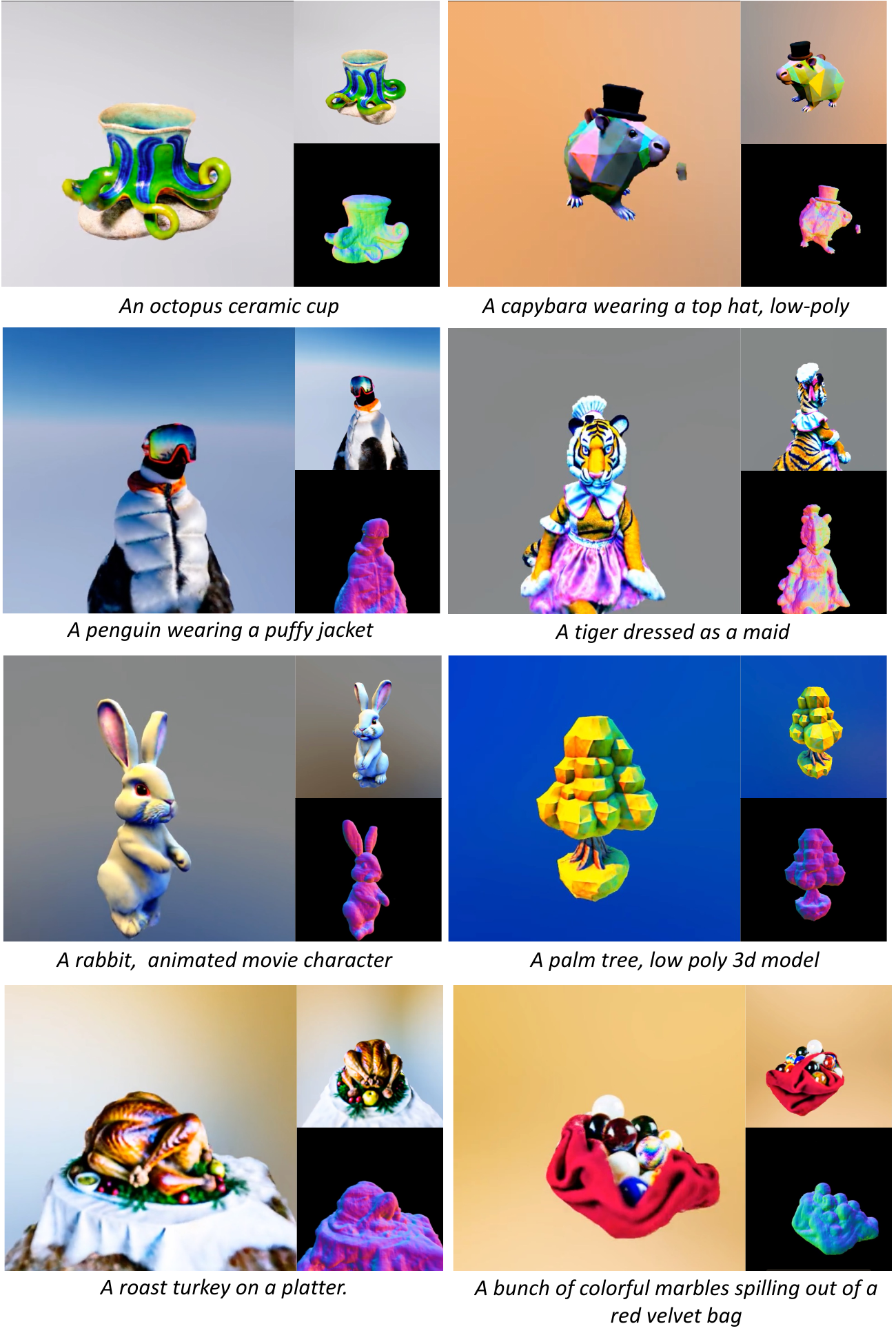}
\caption{More qualitative results on text-to-NeRF generation.}
\label{fig:high_res_nerf2}
\end{figure*}

\begin{figure*}[t]
\centering
\includegraphics[width=0.75\linewidth]{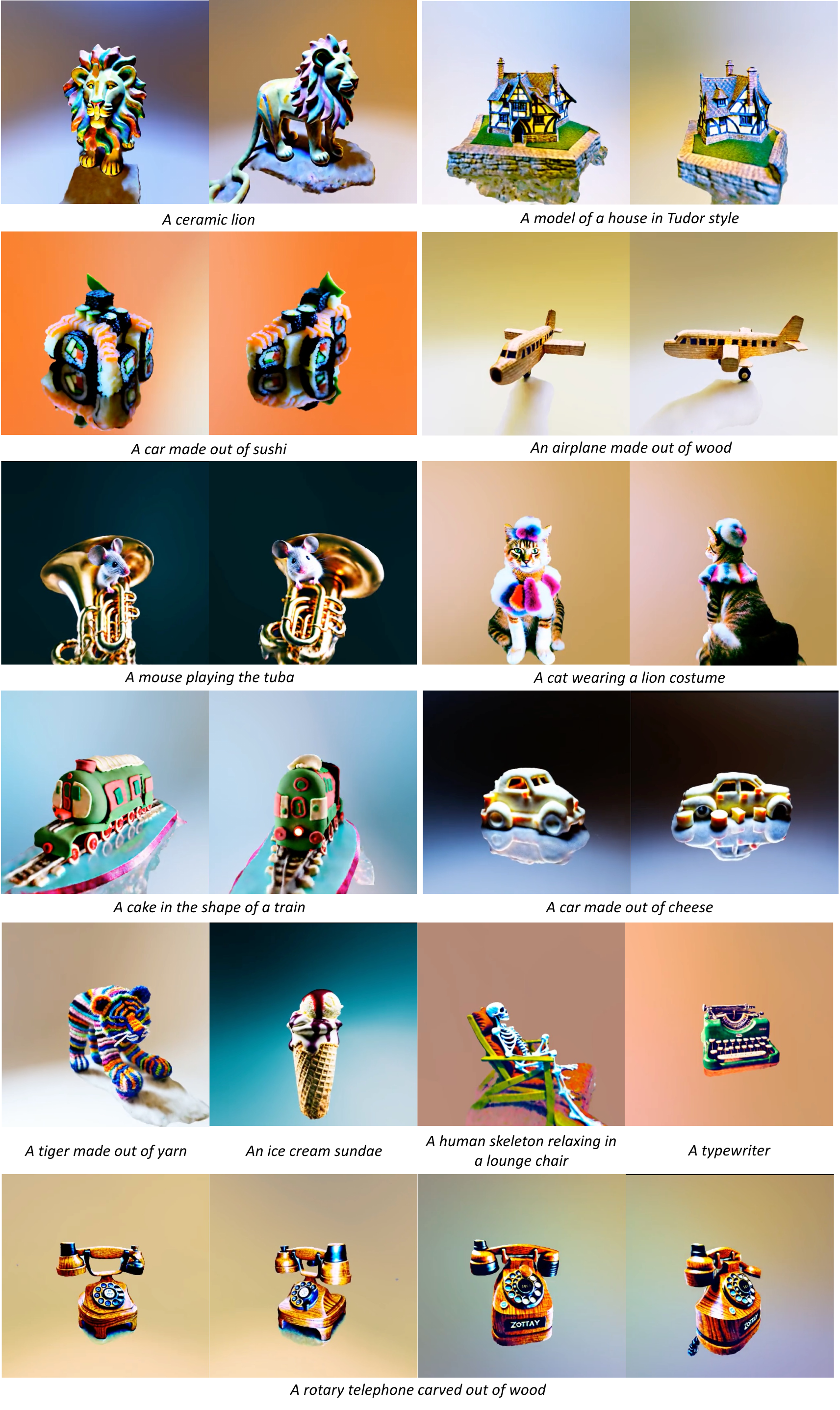}
\caption{More qualitative results on text-to-NeRF generation.}
\label{fig:high_res_nerf3}
\end{figure*}

We present additional 3D generation results with more prompts in \cref{fig:high_res_nerf2} and \cref{fig:high_res_nerf3}. 


\begin{figure*}[t]
\centering
\includegraphics[width=\linewidth]{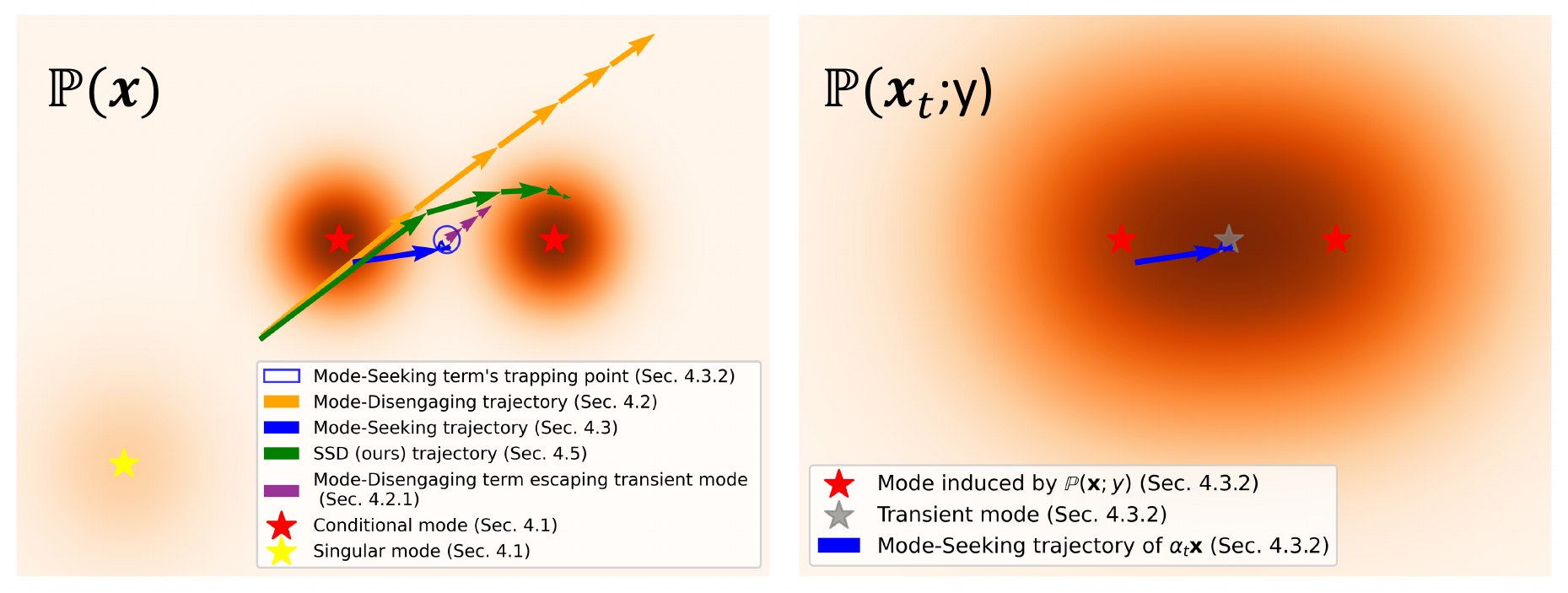}
\vspace{-2.2em}
\caption{\textbf{Evaluating the learning behaviour of $\theta$ supervised by different estimators with an illustrative 2D example.} In this toy example $\theta=\boldsymbol{x}\in \mathbb{R}^2$, and $\mathbb{P}(\boldsymbol{x})$ is a mixture of Gaussian distributions. The setup details are in \cref{sec:toy_example}. We initialize $\theta$ to be at the yellow star, optimize it with \cref{eq:van_sds} under different estimators, and record the learning trajectories of $\theta$. Note that ideally $\theta$ should converge to any modes around the red stars. \textit{Mode-Disengaging trajectory}: the learning trajectory of $\boldsymbol{x}$ by employing the mode-disengaging term only in \cref{eq:van_sds}, where \textit{Mode-Seeking trajectory} and \textit{SSD trajectory} are defined similarly. The purple trajectory describes how $\theta$ evolves when initialized at the trapping point and supervised by mode-disengaging term. On the right, we verify that the mode-seeking term indeed tries to approximate the transient mode for large $t$, causing the convergence point to lie between the normal modes of $\mathbb{P}(\boldsymbol{x})$ on the left. We present the detailed experiment analysis in \cref{sec:toy_example}.} 
\label{fig:intuition}
\vspace{-1.2em}
\end{figure*}

\begin{figure}
    \centering
    \includegraphics[width=0.5\linewidth]{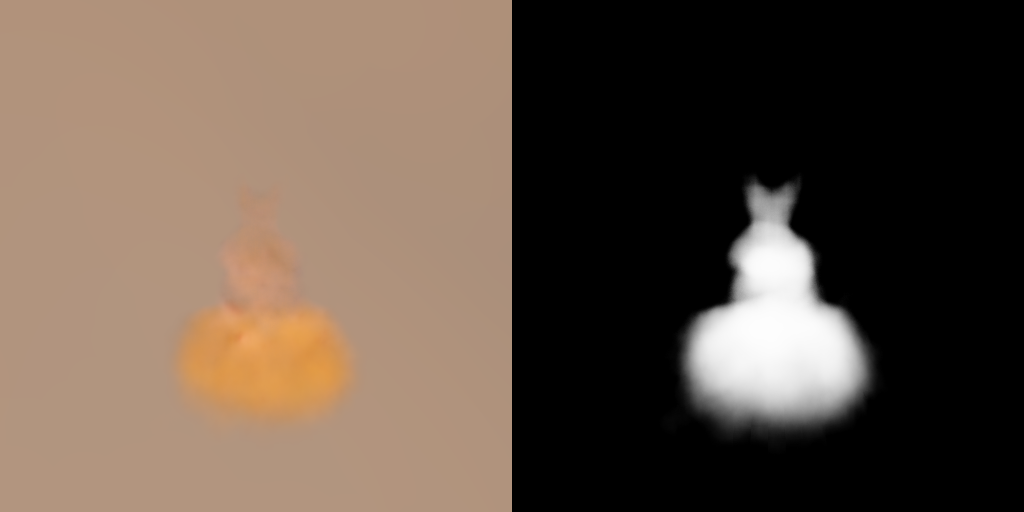}
    \caption{A 3D asset generated by applying the mode-seeking term alone for 15000 steps. The prompt is ``a zoomed out DSLR photo of a baby bunny sitting on top of a pile of pancake''.}
    \label{fig:3d_mode_seeking_only}
\end{figure}

\section{Analysis for Optimal c}\label{sec:proof_close_c}
For ease of mathematical analysis, in this work we adopt \textit{total variance} as our variance definition in \cref{eq:c_min_var} for its wide usage~\cite{wold1987principal}. We define $\Sigma_{XY}$ as the covariance matrix between random vectors $X$ and $Y$, and define $\Sigma_{X}$ as a shorthand for $\Sigma_{XX}$. Then \cref{eq:c_min_var} becomes $c(\boldsymbol{x},t)=\text{argmin}_k\ \text{tr}(\Sigma_{\hat{\boldsymbol{\epsilon}}_\phi - k\boldsymbol{\epsilon}})$, which has a closed-form solution (see \cref{sec:subsec_for_proof_c} for proof):
\begin{equation}
\label{eq:c_final}
    c(\boldsymbol{x},t)=\frac{\text{tr}(\Sigma_{\hat{\boldsymbol{\epsilon}}_\phi \boldsymbol{\epsilon})}}{\text{tr}(\Sigma_{\boldsymbol{\epsilon}})}=\frac{\text{tr}(\Sigma_{\hat{\boldsymbol{\epsilon}}_\phi \boldsymbol{\epsilon})}}{d},
\end{equation}
where we assume $\boldsymbol{\epsilon}\in \mathbb{R}^d$ and $\text{tr}(\cdot)$ denotes the trace of a matrix. Note that evaluating \cref{eq:c_final} requires estimating the covariance between $\hat{\boldsymbol{\epsilon}}_\phi$ and $\boldsymbol{\epsilon}$ with Monte-Carlo, which is computationally expensive. In our experiments, we monitor the projection ratio of $\hat{\boldsymbol{\epsilon}}_\phi$ on $\boldsymbol{\epsilon}$, defined as
\begin{equation}
    r(\boldsymbol{x}_t, \boldsymbol{\epsilon}, y, t)=\frac{\hat{\boldsymbol{\epsilon}}_\phi (\boldsymbol{x}_t,y,t)\cdot \boldsymbol{\epsilon}}{||\boldsymbol{\epsilon}||^2}.
\end{equation}
 Note that $r$ is defined this way to minimize the norm of $\hat{\boldsymbol{\epsilon}}_\phi (\boldsymbol{x}_t,y,t) - k\boldsymbol{\epsilon}$, namely $r(\boldsymbol{x}_t, \boldsymbol{\epsilon}, y,t)=\text{argmin}_k||\hat{\boldsymbol{\epsilon}}_\phi (\boldsymbol{x}_t,y,t) - k\boldsymbol{\epsilon}||.$ Our experiments reveal that $r$ is quite robust to $\boldsymbol{\epsilon}$ for any combination of $\boldsymbol{x}$, $y$, and $t$, and $c$ is tightly bounded by the extremes of $r$, as shown in App. \cref{fig:c_in_r}.

\section{Proof for the Closed Form of Optimal c.}\label{sec:subsec_for_proof_c} 

Recall from \cref{sec:var_red_form} that, 
\begin{equation*}
    \begin{split}
        c(\boldsymbol{x}, t) &= \text{argmin}_k \left[\text{tr}(\Sigma_{\hat{\boldsymbol{\epsilon}_{\phi}} - k \boldsymbol{\epsilon}})\right]\\
        &= \text{argmin}_k \left[\text{tr}(\Sigma_{\hat{\boldsymbol{\epsilon}}_{\phi}}) - 2k \ \text{tr}(\Sigma_{\hat{\boldsymbol{\epsilon}}_{\phi} \boldsymbol{\epsilon}}) + k^2 \text{tr}(\Sigma_{\boldsymbol{\epsilon}})\right]\\
        &= \text{argmin}_k \left[- 2k \ \text{tr}(\Sigma_{\hat{\boldsymbol{\epsilon}}_{\phi} \boldsymbol{\epsilon}}) + k^2 \text{tr}(\Sigma_{\boldsymbol{\epsilon}})\right],\\
    \end{split}
\end{equation*}
taking the derivative of the above quantity with regard to $k$:
\begin{equation*}
    \begin{split}
        \nabla_k \left[k^2 \text{tr}(\Sigma_{\boldsymbol{\epsilon}}) - 2k \ \text{tr}(\Sigma_{\hat{\boldsymbol{\epsilon}}_{\phi} \boldsymbol{\epsilon}})\right] = 2k \text{tr}(\Sigma_{\boldsymbol{\epsilon}}) - 2 \ \text{tr}(\Sigma_{\hat{\boldsymbol{\epsilon}}_{\phi} \boldsymbol{\epsilon}}),
    \end{split}
\end{equation*}
which is equal to zero if and only if $k=\frac{\text{tr}(\Sigma_{\hat{\boldsymbol{\epsilon}}_\phi \boldsymbol{\epsilon}})}{\text{tr}(\boldsymbol{\epsilon})}$. As $\nabla^2_k [k^2 \text{tr}(\Sigma_{\boldsymbol{\epsilon}}) - 2k \ \text{tr}(\Sigma_{\hat{\boldsymbol{\epsilon}}_{\phi} \boldsymbol{\epsilon}})] = 2\text{tr}(\Sigma_\epsilon) > 0$, the function $[k^2 \text{tr}(\Sigma_{\boldsymbol{\epsilon}}) - 2k \ \text{tr}(\Sigma_{\hat{\boldsymbol{\epsilon}}_{\phi} \boldsymbol{\epsilon}})]$ attains its minimum at the critical point $k=\frac{\text{tr}(\Sigma_{\hat{\boldsymbol{\epsilon}}_\phi \boldsymbol{\epsilon}})}{\text{tr}(\boldsymbol{\epsilon})}$. Consequently, the optimal variance-reducing scale $c^{*}(\boldsymbol{x}, t)=\frac{\text{tr}(\Sigma_{\hat{\boldsymbol{\epsilon}}_\phi \boldsymbol{\epsilon}})}{\text{tr}(\Sigma_{\boldsymbol{\epsilon}})}$. Given that $\boldsymbol{\epsilon}\sim \mathcal{N}(0, I)$, $\Sigma_{\boldsymbol{\epsilon}}=I_d$ where $d$ represents the dimension of $\boldsymbol{\epsilon}$, $c^{*}(\boldsymbol{x}, t)$ can be further simplified to $\frac{\text{tr}(\Sigma_{\hat{\boldsymbol{\epsilon}}_\phi \boldsymbol{\epsilon}})}{d}$.

\section{Proof for the $t$ Dependence of Correlation between the Mode-Seeking Term and $\boldsymbol{\epsilon}$.}\label{sec:proof_linear_cor}
Recall that throughout our theoretical analysis, we make a slightly simplified assumption that $\alpha_0 = 1, \sigma_0 = 0$ and $\alpha_T = 0, \sigma_T = 1$. Here we support the statement in \cref{sec:num_mode_seek} that the linear correlation between the mode-seeking term and $\boldsymbol{\epsilon}$ is dependent on timestep $t$.

When $t=0$, we prove that $\nabla_{\boldsymbol{x}_0} \text{log}\mathbb{P}_{\phi}(\boldsymbol{x}_0 ; y)$ is independent of $\boldsymbol{\epsilon}$. Observing that $\boldsymbol{x}_0= \boldsymbol{x} + 0\cdot \boldsymbol{\epsilon}$, it is evident that the inputs to $\nabla_{\boldsymbol{x}_0} \text{log}\mathbb{P}_{\phi}(\boldsymbol{x}_0 ; y)$ do not contain any information about $\boldsymbol{\epsilon}$. Therefore $\nabla_{\boldsymbol{x}_0} \text{log}\mathbb{P}_{\phi}(\boldsymbol{x}_0 ; y)$ is independent of $\boldsymbol{\epsilon}$. 

When $t=T$, we prove that $\nabla_{\boldsymbol{x}_T} \text{log}\mathbb{P}_{\phi}(\boldsymbol{x}_T ; y)$ is collinear to $\boldsymbol{\epsilon}$:\\
\begin{equation}
    \begin{split}
        \nabla_{\boldsymbol{x}_T} \text{log}\mathbb{P}_{\phi}(\boldsymbol{x}_T ; y) &= \nabla_{\boldsymbol{x}_T} \text{log}\mathcal{N}(\boldsymbol{x}_T; 0, I)\\
        &=  \nabla_{\boldsymbol{x}_T}\text{log}( (2\pi)^{-\frac{d}{2}} exp(-\frac{1}{2}\boldsymbol{x}_T^T \boldsymbol{x}_T))\\
        &=  \nabla_{\boldsymbol{x}_T} (-\frac{1}{2}\boldsymbol{x}_T^T \boldsymbol{x}_T)\\
        &= -\frac{1}{2} \nabla_{\boldsymbol{x}_T} \boldsymbol{x}_T^T \boldsymbol{x}_T\\
        &= -\boldsymbol{x}_T\\
        &= -0\boldsymbol{x}-1\boldsymbol{\epsilon}\\
        &= -\boldsymbol{\epsilon}\\
    \end{split}
\end{equation}
where the superscript $T$ for $\boldsymbol{x}_T^T$ represents vector transpose. 

\section{More Numerical Experiments}\label{sec:numerical_exp}
\begin{figure}
    \centering
    \includegraphics[width=0.5\linewidth]{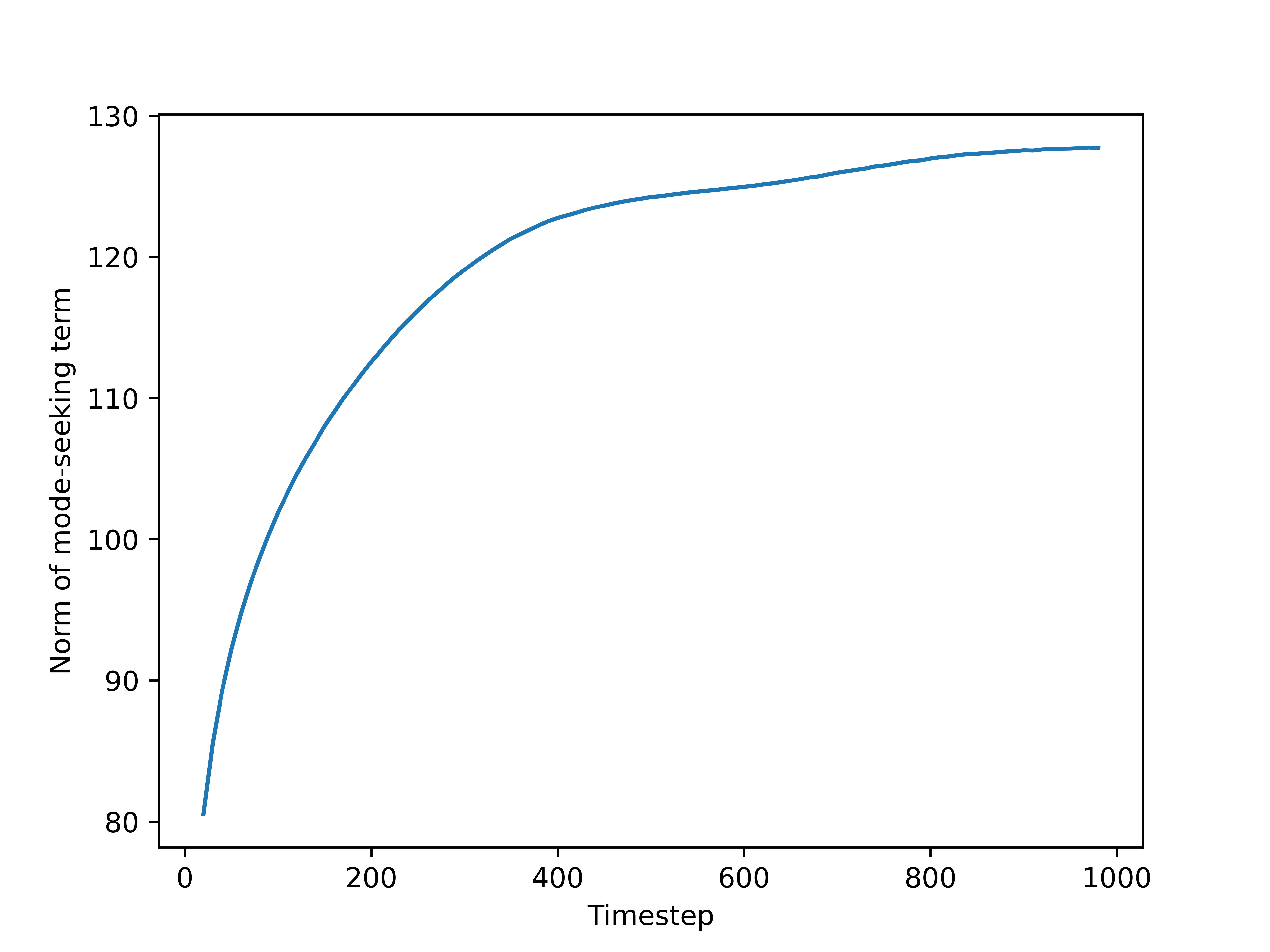}
    \caption{$||\hat{\boldsymbol{\epsilon}}_{\phi}||$ increases as diffusion timestep $t$ increases. This together with \cref{fig:linear_cor} justifies our design of adaptive variance reduction. See \cref{sec:numerical_exp} for the data collection protocol.}
    \label{fig:scale_mode_seeking}
\end{figure}

\begin{figure}
    \centering
    \includegraphics[width=0.5\linewidth]{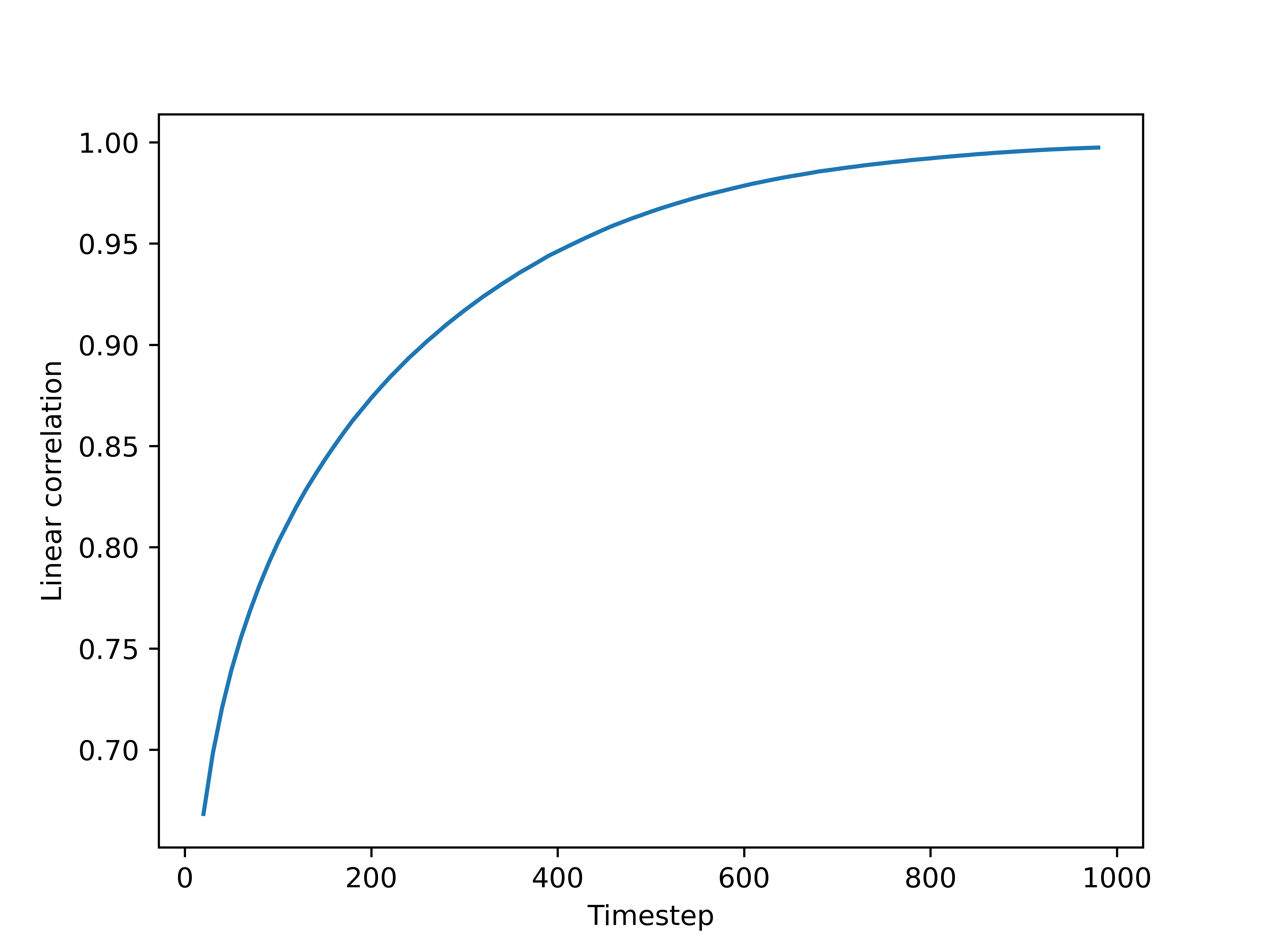}
    \caption{Linear correlation~\cite{puccetti2022measuring} between $\hat{\boldsymbol{\epsilon}}_\phi$ and $\boldsymbol{\epsilon}$ increases with diffusion timestep $t$. This together with \cref{fig:scale_mode_seeking} justifies our design of adaptive variance reduction. See \cref{sec:numerical_exp} for the data collection protocol.}
    \label{fig:linear_cor}
\end{figure}

\begin{figure}
    \centering
    \includegraphics[width=0.5\linewidth]{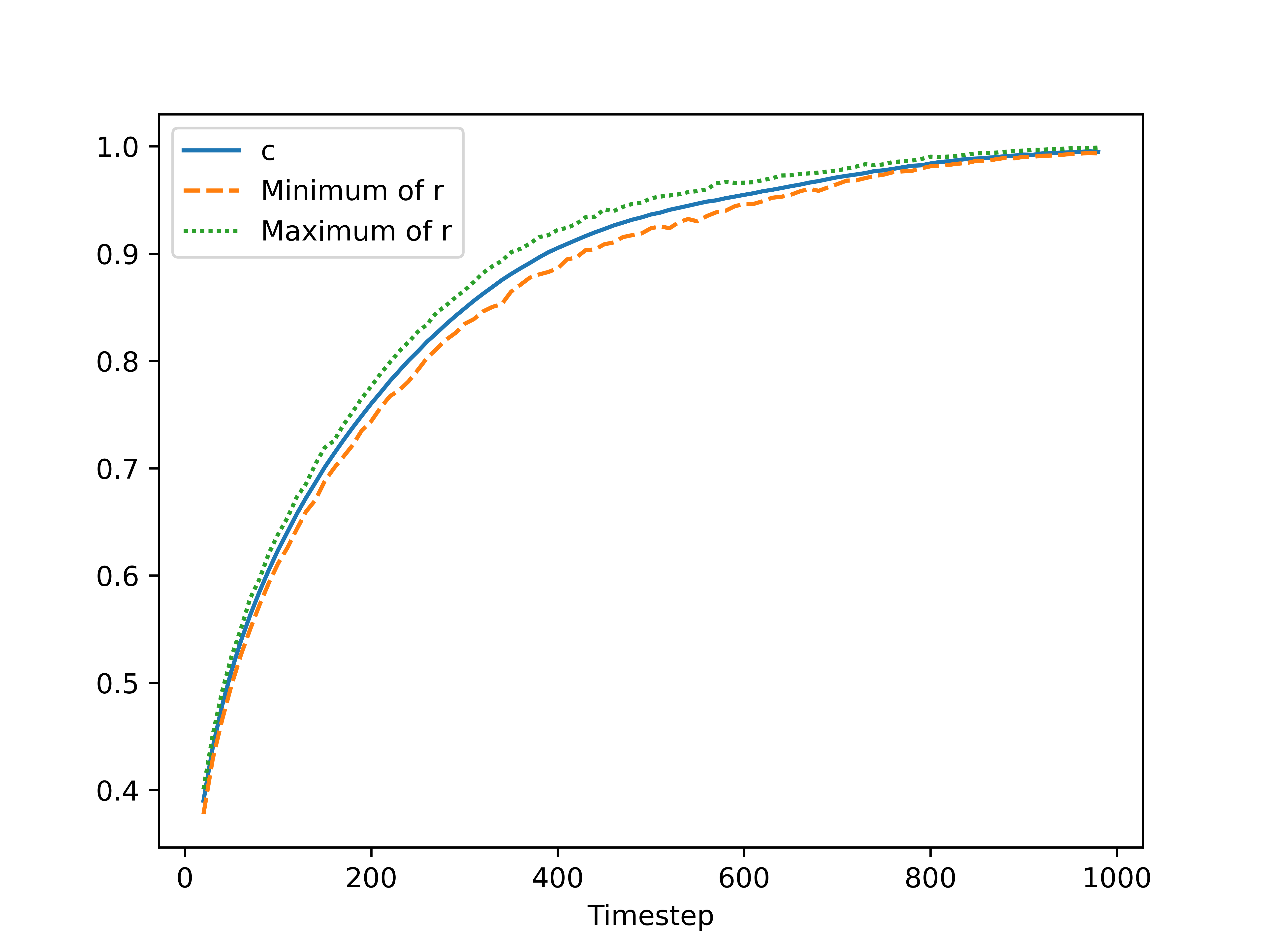}
    \caption{The optimal $c(\boldsymbol{x}, t)$ lies within the extremes of our proposed proxy $r$. For each timestep, we compute $r$ as defined in \cref{eq:def_r} and $c$ as defined in \cref{eq:c_final}. Note that with the 8192 samples as discussed in \cref{sec:numerical_exp}, $c$ is unique while $r$ can take 8192 different values. Therefore the maximum and minimum of these $r$ are visualized separately. It is evident that $r$ and $c$ are numerically close with similar trend, validating that $r$ is a feasible proxy for $c$. See \cref{sec:numerical_exp} for the data collection protocol.}
    \label{fig:c_in_r}
\end{figure}

\begin{figure}
    \centering
    \includegraphics[width=0.5\linewidth]{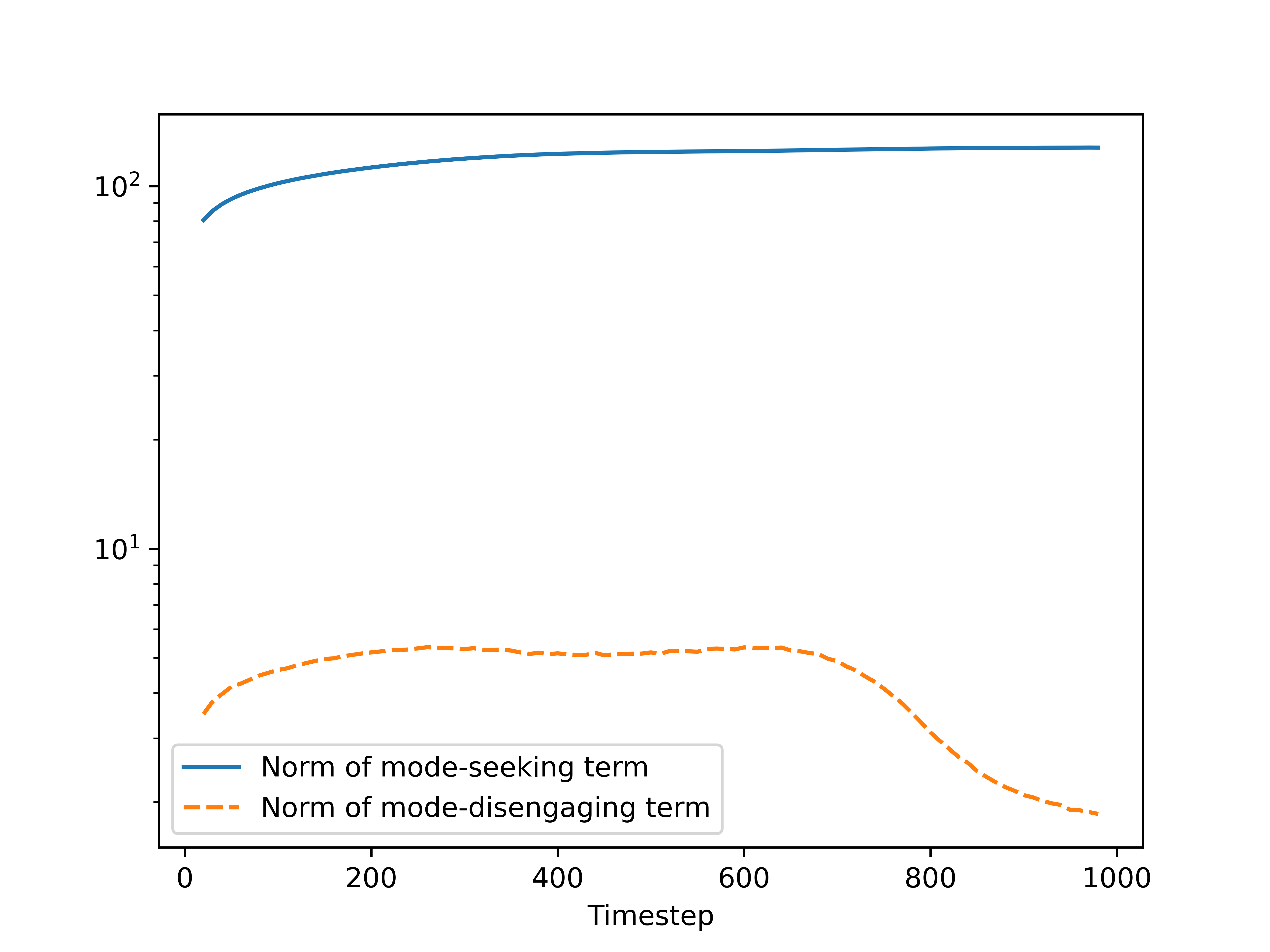}
    \caption{Comparison between norms of the mode-seeking and mode-disengaging terms. Note that the y-axis is plotted in log scale. $||\hat{\boldsymbol{\epsilon}}_\phi|| \in [80, 130]$ while $||h||\in [2, 5.5]$. The vast norm difference between the two terms constitutes an important justification for our estimator-rescaling in the proposed stable score distillation (SSD) \cref{alg:ours} (L15). See \cref{sec:numerical_exp} for the data collection protocol.}
    \label{fig:norm_of_h_and_mode_seeking}
\end{figure}

Numerical experiments are conducted to illustrate and validate the properties of supervision signals employed for 3D generation, as well as the motivation and correctness of our method details. We first sample 2D renderings of SDS-trained 3D assets. Then for each rendering and each time step $t$, 8192 noises $\boldsymbol{\epsilon}$ are randomly sampled, resulting in 8192 $\boldsymbol{x}_t$. We compute $r$, $c$, $||h||$ and other relevant quantities based on the sampled noises and $\boldsymbol{x}_t$, which are visualized in \cref{fig:scale_mode_seeking},  \cref{fig:linear_cor}, \cref{fig:c_in_r} and \cref{fig:norm_of_h_and_mode_seeking}.






\section{Connection with Common Observations and Practices in 3D Content Generation}\label{sec:connection}
\begin{figure}
    \centering
    \includegraphics[width=0.45\textwidth]{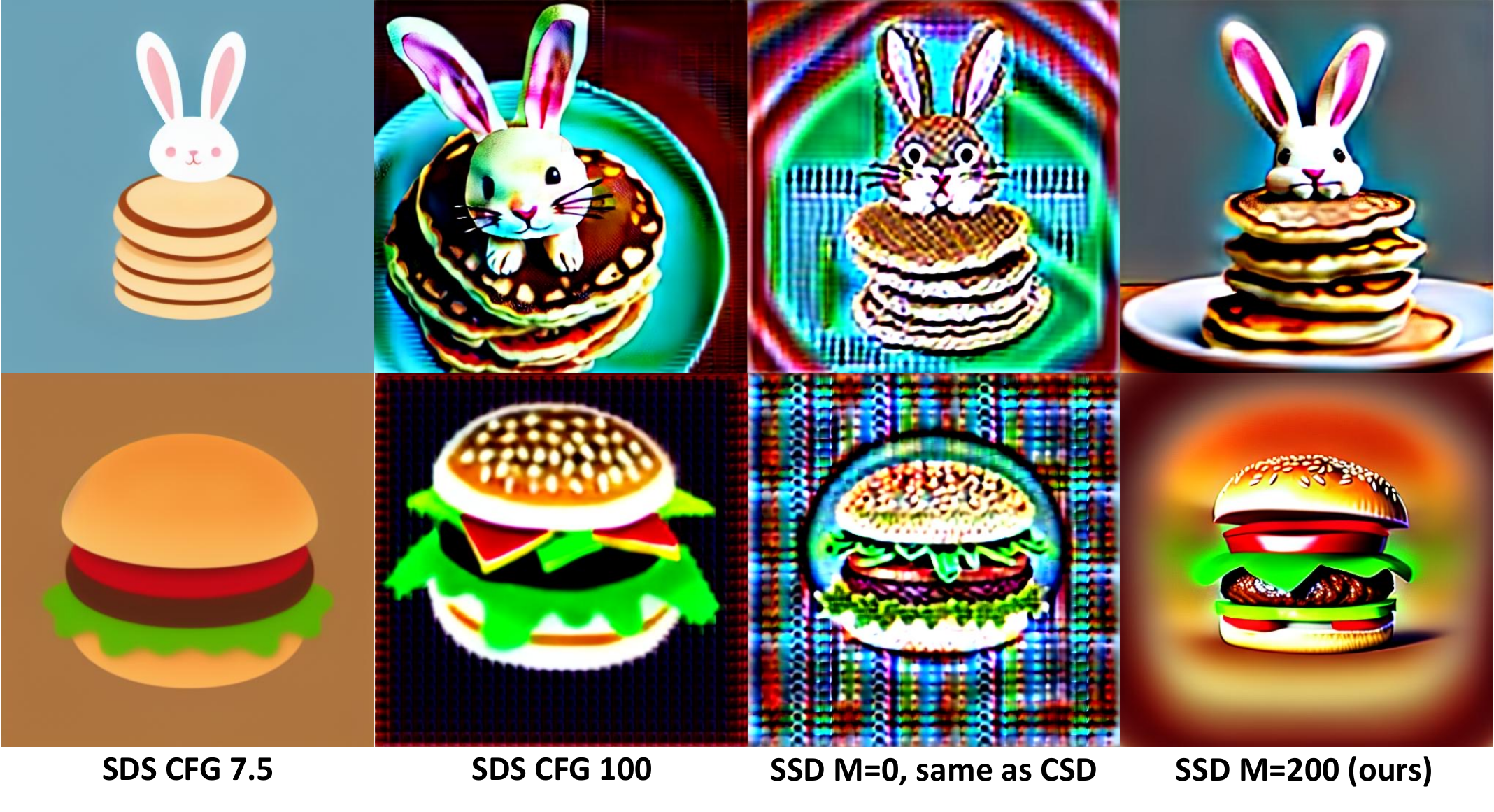}
    \vspace{-1em}
    \caption{\textbf{Comparisons on 2D optimization.} The experiments are run with more training iterations than normally used to expose the intrinsic properties of the estimators. Objects' local features are more plausible with SSD. It is evident that while the mode-seeking term contributes to finer local details, it also incurs color implausibility.}
    \label{fig:2d_comparison}
    \vspace{-1.5em}
\end{figure}

\begin{itemize}
    \item \textbf{Large CFG Scales.} As illustrated in App. \cref{fig:norm_of_h_and_mode_seeking}, the norms of the mode-disengaging term and the mode-seeking term are vastly different. Consequently, a large CFG scale is necessary to make the scale of the main learning signal $\omega \cdot h$ at least comparable to that of the mode-seeking term. Otherwise the \colorbox{mygred}{SDS estimator} in \cref{eq:van_sds} would be dominated by the mode-seeking term, which can only generate over-smoothed contents for reasons discussed in \cref{sec:over_smoothing}, as visualized in \cref{fig:3d_mode_seeking_only}.
    \item \textbf{Over-smoothness.} In SDS where $\omega=100$, $||\hat{\boldsymbol{\epsilon}}_\phi||$ and $||\omega h||$ are on the same order of magnitude, and the mode-disengaging term lacks the dominance to the SDS estimator necessary to mitigate the over-smoothing effect induced by the mode-seeking term. 
    \item \textbf{Color Saturation.} This challenge has been repeatedly observed in text-to-3D generation~\citep{poole2022dreamfusion}. Recent work of DreamTime~\cite{huang2023dreamtime} also observes severe color saturation tendency when $t<100$. According to our analysis in \cref{sec:ana_mode_dis}, when $t$ is small the mode-disengaging term tends to drive the rendering $g(\theta, c)$ away from modes of the natural image distribution, making the rendering implausible. As pointed out in \cite{choi2022perception}, in practice the diffusion model tends to influence fine details and colors, but not large-scale geometry, when $t$ is small. Thus with small $t$ the mode-disengaging term generates implausible colors, without corrupting the general geometry of the objects. We provide experiments in \cref{fig:2d_comparison} to show that the mode-disengaging term actually causes color saturation with 2D experiments.
    
\end{itemize}

\section{Illustrative Example}\label{sec:toy_example}
We evaluate the learning behaviour of the mode-seeking, mode-disengaging and SSD terms with an illustrative example, as shown in \cref{fig:intuition}. In this experiment, we assume that $\boldsymbol{x} = \theta\in \mathbb{R}^2$, and $\mathbb{P}_{\phi}(\boldsymbol{x})=0.2\mathcal{N}([0,0]^T, 0.1I)+0.4\mathcal{N}([1,1]^T, 0.05I)+0.4\mathcal{N}([2,1]^T, 0.05I)$ is a mixture of Gaussian distributions. The two Gaussian components located at $[1,1]^T$ and $[2,1]^T$ are conditional modes, while the one at $[0,0]^T$ is a singular one. Note that we want $\theta$ to approximate the two conditional modes as close as possible. We initialize $\theta$ at $[0,0]^T$ to simulate the situation that all 3D generation algorithms begin with empty renderings.
We then substitute the mode-seeking, mode-disengaging and our SSD estimators for the SDS estimator in \cref{eq:van_sds} one by one, and record the learning trajectory of $\theta$. To validate the trap-escaping property of the mode-disengaging term we initialize a $\theta$ to the trapping point, namely $[1.5, 1]^T$, and use $h$ to supervise $\theta$ afterwards. The learning trajectories are illustrated in \cref{fig:intuition} (left). 
We observe that the mode-seeking trajectory gets trapped between the two conditional modes rapidly, not converging to any specific conditional mode of $\mathbb{P}_{\phi}(\boldsymbol{x})$. To inspect whether the issue is caused by the transient-mode problem, we present the density map of $\mathbb{P}_{\phi}(\boldsymbol{x}_t; y)$ at $t=350$, and the learning trajectory of $\alpha_t \boldsymbol{x}$ in \cref{fig:intuition} (right). The density map reveals that the two conditional modes of $\mathbb{P}_{\phi}(\boldsymbol{x};y)$ result in a transient mode in $\mathbb{P}_{\phi}(\boldsymbol{x}_t;y)$, whose probability density is higher than the induced modes. And $\alpha_t \boldsymbol{x}_t$ indeed approaches the transient mode $\alpha_t \boldsymbol{o}_{tr}$. 
Conversely, the mode-disengaging term can propel $\theta$ away from the trapping point $\boldsymbol{o}_{tr}$. However, although at the beginning it guides $\theta$ towards a conditional mode, it ultimately steers $\theta$ into a low-density region of $\mathbb{P}_{\phi}(\boldsymbol{x})$. In contrast, the proposed SSD swiftly guides $\theta$ towards a conditional mode at beginning, avoids getting trapped, and finally converges to a point with high density in $\mathbb{P}_\phi(\boldsymbol{x})$.

\end{document}